\PassOptionsToPackage{dvipsnames, table}{xcolor}
\documentclass[letterpaper]{nature}

\usepackage{color}
\usepackage{xurl}
\usepackage{verbatim}
\usepackage{multirow}
\usepackage{xspace}
\usepackage{graphicx}
\usepackage{amsmath} 
\usepackage{bbm}
\usepackage{amssymb}
\usepackage{booktabs} 
\usepackage{times} 
\usepackage{lscape}
\usepackage[left=1in,right=1in,bottom=1.1in,top=1in]{geometry}
\usepackage{enumitem}
\usepackage[table,xcdraw]{xcolor}
\usepackage[normalem]{ulem}
\usepackage{xcolor}
\usepackage[innercaption]{sidecap}
\usepackage{lineno}
\usepackage{longtable}
\usepackage{tcolorbox}

\usepackage[dvipsnames]{xcolor}
\usepackage[framemethod=tikz]{mdframed} 
\usepackage[table]{xcolor}
\mdfdefinestyle{promptsllm}{
	innertopmargin=1.2\baselineskip,
	innerbottommargin=0.8\baselineskip,
	roundcorner=5pt,
	nobreak,
	singleextra={%
		\draw(P-|O)node[xshift=1em,anchor=west,fill=Yellow!15,draw,rounded corners=5pt]{%
		  \promptVanillaTitle};
	},
}
\newenvironment{promptsllm}[1][Prompts for Evaluating LLMs]{
	\bigskip
	\gdef\promptVanillaTitle{#1}
	\begin{mdframed}[style=promptsllm, backgroundcolor=white]
}{
	\end{mdframed}
	\medskip
}

\usepackage{xr-hyper}
\makeatletter
\newcommand*{\addFileDependency}[1]{
  \typeout{(#1)}
  \@addtofilelist{#1}
  \IfFileExists{#1}{}{\typeout{No file #1.}}
}
\makeatother

\usepackage[colorlinks,citecolor=blue,urlcolor=blue]{hyperref}
\usepackage[margin=-1cm, labelfont=bf, font=footnotesize]{caption}
\usepackage[ruled,vlined]{algorithm2e}
\SetKwRepeat{Do}{do}{while}

\newcommand{\xhdr}[1]{\vspace{1mm}\noindent{{\bf #1}}}

\newcommand{\hide}[1]{}

\let\oldnl\nl
\newcommand{\nonl}{\renewcommand{\nl}{\let\nl\oldnl}}  
\usepackage{bm}
\graphicspath{{./FIG/}}

\newcommand{\name}{\textsc{TxAgent}\xspace} 
 
\newcommand{\toolgen}{\textsc{ToolGen}\xspace} 
\newcommand{\questgen}{\textsc{QuestionGen}\xspace} 
\newcommand{\tracegen}{\textsc{TraceGen}\xspace} 
\newcommand{\toolrag}{\textsc{ToolRAG} model\xspace} 
\newcommand{\toolbox}{\textsc{ToolUniverse}\xspace} 
\newcommand{\trainset}{\textsc{TxAgent-Instruct}\xspace}
\newcommand{\onlineref}[1]{Online Methods Section~\ref{#1}\xspace} 

\newcommand{\extfigref}[1]{Extended Data Figure~\ref{#1}}

\title{\begin{center}
TxAgent: An AI Agent for Therapeutic Reasoning \\ Across a Universe of Tools
\end{center}\vspace{-10mm}}    

\author  
{\small\begin{center}   
Shanghua Gao$^{1}$, Richard Zhu$^{1}$, Zhenglun Kong$^{1}$, Ayush Noori$^{1}$, Xiaorui Su$^{1}$, Curtis Ginder$^{1,2}$, \\ Theodoros Tsiligkaridis$^{3}$, and Marinka Zitnik$^{1,4,5,6,\ddag}$ \\[2mm]  
{$^{1}$Department of Biomedical Informatics, Harvard Medical School, Boston, MA} \\
{$^{2}$Cardiovascular Division, Department of Medicine, Brigham and Women’s Hospital, Harvard Medical School, Boston, MA} \\
{$^{3}$MIT Lincoln Laboratory,  Lexington, MA} \\
{$^{4}$Kempner Institute for the Study of Natural and Artificial Intelligence, Harvard University, Cambridge, MA} \\
{$^{5}$Broad Institute of MIT and Harvard, Cambridge, MA} \\
{$^{6}$Harvard Data Science Initiative, Cambridge, MA} \\
{$\ddag$Corresponding author. Email: marinka@hms.harvard.edu}\\[2mm]
\name project is at \url{https://zitniklab.hms.harvard.edu/TxAgent}\\
\name code and demos are at \url{https://github.com/mims-harvard/TxAgent}\\
\toolbox is at \url{https://github.com/mims-harvard/ToolUniverse}
\end{center}
}

\begin{document}

\maketitle

{\spacing{1.15}
\begin{abstract}

Precision therapeutics require multimodal adaptive models that generate personalized treatment recommendations. We introduce \name, an AI agent that leverages multi-step reasoning and real-time biomedical knowledge retrieval across a toolbox of 211 tools to analyze drug interactions, contraindications, and patient-specific treatment strategies. \name evaluates how drugs interact at molecular, pharmacokinetic, and clinical levels, identifies contraindications based on patient comorbidities and concurrent medications, and tailors treatment strategies to individual patient characteristics, including age, genetic factors, and disease progression.
\name retrieves and synthesizes evidence from multiple biomedical sources, assesses interactions between drugs and patient conditions, and refines treatment recommendations through iterative reasoning. It selects tools based on task objectives and executes structured function calls to solve therapeutic tasks that require clinical reasoning and cross-source validation. The \toolbox consolidates 211 tools from trusted sources, including all US FDA-approved drugs since 1939 and validated clinical insights from Open Targets.
\name outperforms leading LLMs, tool-use models, and reasoning agents across five new benchmarks: DrugPC, BrandPC, GenericPC, TreatmentPC, and DescriptionPC, covering 3,168 drug reasoning tasks and 456 personalized treatment scenarios. It achieves 92.1\% accuracy in open-ended drug reasoning tasks, surpassing GPT-4o by up to 25.8\% and outperforming DeepSeek-R1 (671B) in structured multi-step reasoning. \name generalizes across drug name variants and descriptions, maintaining a variance of $<$0.01 between brand, generic, and description-based drug references, exceeding existing tool-use LLMs by over 55\%.
By integrating multi-step inference, real-time knowledge grounding, and tool-assisted decision-making, \name ensures that treatment recommendations align with established clinical guidelines and real-world evidence, reducing the risk of adverse events and improving therapeutic decision-making.
\end{abstract}
}

\clearpage

\spacing{1.38}

\section*{Main}


Precision therapy personalizes treatments based on individual patient conditions to maximize efficacy and minimize risks. Prescribing the appropriate drug requires evaluating multiple factors, including patient-specific characteristics, comorbidities, drug interactions, contraindications, current clinical guidelines, drug mechanisms of action, and the underlying biology of the disease~\cite{huang2024foundation}.
Large language models (LLMs) can process therapeutic tasks by large-scale pretraining~\cite{dubey2024llama,radford2018improving,jiang2023mistral,touvron2023llama,bai2023qwen} followed by fine-tuning on medical data~\cite{singhal2023large,singhal2025toward,chen2023meditron}. While LLMs generate fluent, contextually relevant responses, they lack real-time access to updated biomedical knowledge, frequently hallucinate, and cannot reliably reason over multiple clinical variables. Retraining these models with new medical insights is computationally expensive and impractical due to catastrophic forgetting. Furthermore, LLMs absorb large volumes of open-net data, which may contain unverified or deliberately misleading medical information~\cite{alber2025medical}.
Tool-augmented LLMs~\cite{berkeley-function-calling-leaderboard,watt-tool-8B,liu2024toolace} incorporate external knowledge retrieval mechanisms, such as retrieval-augmented generation (RAG)~\cite{gao2023retrieval}, to mitigate these issues. These models retrieve drug and disease information from external sources but cannot execute multi-step reasoning required for treatment selection. Precision therapy could benefit from iterative reasoning, where models could retrieve information from verified sources, evaluate interactions, and dynamically refine treatment plans.

We introduce \name, an AI agent~\cite{gao2024agent,boiko2023autonomous,yao2023react,bran2023chemcrow,swanson2024virtual} that delivers evidence-grounded treatment recommendations by combining multi-step reasoning with real-time biomedical tool integration. \name generates natural language responses alongside a transparent reasoning trace, detailing each step of its decision-making process. It executes goal-driven tool selection, calling external databases and specialized machine learning (ML) models to ensure accuracy. To support complex medical queries, \name leverages \toolbox, a biomedical toolbox consolidating 211 expert-curated tools, spanning drug mechanisms, interactions, clinical guidelines, and disease annotations. These tools integrate trusted sources, including openFDA\cite{kass2016openfda}, Open Targets\cite{ochoa2023next}, and the Human Phenotype Ontology~\cite{castellanos2024human}. \name further employs \toolrag, an ML-based retrieval system that dynamically selects the most relevant tools from \toolbox based on query context.

\name consists of: (1) \toolbox, a diverse collection of 211 biomedical tools, (2) a specialized LLM fine-tuned for multi-step reasoning and tool execution, and (3) \toolrag, an adaptive tool retrieval model. To construct tools compatible with \name, we introduce \toolgen, a multi-agent tool construction system that generates tools from API documentation. \name is fine-tuned using \trainset, a dataset of 378,027 instruction-tuning samples that is derived from 85,340 multi-step reasoning traces and encompasses 177,626 reasoning steps and 281,695 function calls. The dataset is generated using \questgen and \tracegen, multi-agent systems that construct diverse therapeutic queries and generate stepwise reasoning traces that cover treatment and drug information in FDA labels since 1939.

We introduce five new benchmarks (DrugPC, BrandPC, GenericPC, DescriptionPC, TreatmentPC, Table~\ref{table:fda_11tasks}). These benchmarks comprehensively assess drug selection, treatment personalization, and reasoning robustness across structured and unstructured queries. \name outperforms larger LLMs and existing tool-use models across all five benchmarks, achieving state-of-the-art performance in open-ended drug reasoning and patient-specific therapeutic decision-making. On the DrugPC benchmark, which evaluates 11 common drug reasoning tasks, \name attains 92.1\% accuracy in the open-ended setting, where the model generates answers without predefined choices. This performance surpasses GPT-4o\cite{openai2024gpt4o}, the strongest closed-weight reference model, by 25.8\% (GPT-4o: 66.3\%) and outperforms Llama-3.1-70B-Instruct\cite{dubey2024llama}, a model nearly 9× larger, by 39.3\% (Llama-3.1-70B-Instruct: 52.8\%). \name, based on the fine-tuned 8-billion parameter Llama-3.1-8B-Instruct model~\cite{dubey2024llama}, delivers superior accuracy while maintaining computational efficiency. Compared to tool-use LLMs with function-calling capabilities, such as ToolACE and WattTool~\cite{watt-tool-8B,liu2024toolace}, \name significantly outperforms both models in open-ended drug reasoning tasks. Unlike existing tool-augmented LLMs, which struggle with multi-step tool selection and iterative reasoning, \name dynamically retrieves and synthesizes knowledge from 211 biomedical tools, achieving more accurate and context-aware therapeutic decisions.

Beyond drug reasoning, \name generalizes across drug name variants and descriptions, overcoming a key limitation of LLM-based methods~\cite{gallifant2024language,chen2024clinicalbench}. Many models exhibit high variance when drugs are referenced by brand names, generic names, or detailed descriptions~\cite{gallifant2024language}. In contrast, \name achieves an exceptionally low accuracy variance of $<$0.01 across these variations, whereas GPT-4o exhibits a variance of 9.96, indicating a much higher sensitivity to representation shifts. On DescriptionPC, a benchmark that evaluates drug reasoning when drug names are replaced with descriptive narratives, \name attains 56.5\% accuracy, outperforming GPT-4o by 8.3\% and indicating \name's robustness to infer drug identities from contextual clues. \name also excels in personalized treatment recommendations, where it evaluates patient-specific drug selection. On TreatmentPC, which assesses 456 real-world treatment scenarios, \name outperforms GPT-4o by 13.6\% and Llama-3.1-70B-Instruct by 25.4\% in the open-ended setting, establishing its superiority in personalized medicine. Compared to DeepSeek-R1~\cite{guo2025deepseek}, a 671-billion parameter model optimized for multi-step reasoning, \name achieves 7.5\% higher accuracy in open-ended queries, demonstrating that specialized reasoning and tool-use capabilities outweigh model size.

We conduct ablation studies to evaluate \name's toolbox size, tool dependency, and reasoning process. Increasing the number of tools in \toolbox improves performance, demonstrating that access to external biomedical tools improves therapeutic reasoning. We compare real-world tool usage to an LLM acting as a tool substitute and find that tool-assisted decision-making consistently outperforms LLM-only reasoning, highlighting the need for grounding AI agents in continually updated and verified therapeutic knowledge. We also examine the impact of explicit reasoning steps before function calls and show that structured reasoning improves performance more than multi-round function calls alone. Finally, we analyze the effect of multi-step training traces and find that increasing the number of reasoning steps in fine-tuning and inference significantly improves \name's ability to handle complex drug reasoning and treatment selection.

\section*{Results}


\subsection*{\name: Multi-step therapeutic reasoning with a universe of tools}
\name uses multi-step, white-box reasoning and tool-use for solving precision treatment problems (Figure~\ref{fig:fig1}a). Using a wide array of tools that connect to verified knowledge bases, such as FDA-approved drug labels and the Open Targets~\cite{kass2016openfda,ochoa2023next}, as well as machine learning tools for special purposes such as tool retrieval (Figure~\ref{fig:fig1}b), \name performs detailed reasoning on drugs, diseases, and patient populations. This ability to leverage a vast array of biomedical tools ensures \name is not limited by the internal knowledge of LLMs, enabling it to generate accurate and reliable answers with transparent reasoning traces. It can handle a variety of patient scenarios, from specific patient populations and complex medical histories to polypharmacy and individual-specific genetic variants. \name uses \toolbox, which is a generalizable toolbox with 211 tools that support real-time retrieval of knowledge from verified data sources, including openFDA~\cite{kass2016openfda}, Open Targets~\cite{ochoa2023next}, and the Human Phenotype Ontology from the Monarch Initiative~\cite{castellanos2024human}. These tools address diverse aspects of drugs and diseases, such as drug indications and usage (Figure~\ref{fig:fig1}c).

\name is an LLM trained to use tools. This is achieved by building three training datasets (a tooling dataset, a comprehensive therapeutic question dataset, and a reasoning trace dataset), which we create using three auxiliary agent systems (Figure~\ref{fig:fig2}a). Given these datasets, we instruction-tune an LLM~\cite{dubey2024llama} to achieve multiple capabilities, including multi-step reasoning and tool call argument generation. For each step in the multi-step reasoning process, \name receives either a therapeutic question or tool feedback from the previous round. Based on this input, \name generates a language-based thought process and invokes calls to tools. During the reasoning process, to identify and utilize relevant tools, \name invokes the \toolrag, which selects suitable candidates from \toolbox based on descriptions provided by \name. This iterative process continues until \name arrives at a final answer and invokes the \textsc{Finish} tool to conclude the reasoning process. The output of \name includes both the final answer and a multi-step reasoning trace. Each step of the reasoning trace includes a thought process, function calls to utilize tools, and feedback from those tools.
We show the detailed inference process of \name in~\onlineref{sec:skill_txagent} and Algorithm~\ref{alg:txagent_inference}.

\subsection*{Capabilities of \name}

\name generates reasoning traces, constructs function call arguments, performs multi-step logical reasoning, and searches for, selects, and invokes tools to solve a therapeutic reasoning task. These capabilities are developed through instruction tuning of the LLM (\onlineref{sec:skill_txagent}). By applying these capabilities, \name retrieves verified biomedical knowledge through tool calls, selects tools based on specific objectives, solves problems through multi-step reasoning, and integrates continuously updated knowledge bases.

\xhdr{Knowledge grounding using tool calls.}
Treatment decisions require reliable answers with transparent justifications. LLMs lack inherent mechanisms to verify their predictions, requiring users to assess trustworthiness manually. \name addresses this by retrieving verified information from trusted sources through function calls. Instead of generating responses directly, \name queries tools to obtain accurate data and formulates answers based on verified outputs. In Figure~\ref{fig:fig1}f, \name determines the dosage of Kisunla (donanemab-azbt), an FDA-approved drug from 2024, which is beyond the training data of its base LLMs. \name recognizes the knowledge gap, calls \textit{get\_dosage}, and retrieves dosage details from FDA records. It then synthesizes the retrieved information into a response. This approach ensures factual accuracy and transparency, allowing users to verify responses through reasoning traces.

\xhdr{Goal-oriented tool selection.}
\name uses \toolrag to search for, identify, and apply the most relevant tools. Figure~\ref{fig:fig1}g shows \name retrieving adverse reactions for Alyftrek (vanzacaftor, tezacaftor, deutivacaftor). It recognizes the need for external data, generates function call arguments, and queries \toolbox. From the returned tools, \name selects \textit{get\_adverse\_reactions} to extract relevant information from FDA drug labels. This process enables \name to dynamically integrate new tools rather than relying on static, pre-trained knowledge. By first generating a plan and then selecting appropriate tools, \name supports adaptive decision-making.

\xhdr{Multi-step therapeutic reasoning.}
\name applies multi-step reasoning to address complex problems that require integrating multiple sources of information or adapting to incomplete data. Single-step approaches fail when problems demand information from multiple perspectives or when function calls return insufficient results. By iteratively generating reasoning steps and function calls, \name refines its analysis until it reaches a well-supported answer. In Figure~\ref{fig:fig1}h, \name identifies protein targets for breast cancer, a task no single \toolbox tool can complete. Therefore, \name first retrieves the disease’s EFO ID using \textit{get\_disease\_id\_desc}, then queries \toolbox for tools that map diseases to protein targets. From the returned options, \name selects \textit{get\_associated\_targets} and ranks the retrieved proteins by score. This iterative process ensures robust reasoning in cases where direct retrieval is insufficient.

\xhdr{Real-time retrieval from continually updated knowledge sources.}
LLMs retain only the knowledge available at the time of training and cannot update dynamically. Retraining models to incorporate new biomedical information is computationally expensive and impractical. Retrieval-augmented generation \cite{lewis2020retrieval} mitigates this by querying a precomputed vector database, but maintaining high-quality embeddings for frequent updates is resource-intensive. \name addresses this limitation by executing function calls to directly query real-time data sources, such as Open Targets and FDA databases. This approach enables \name to retrieve current drug approvals, clinical guidelines, and treatment indications without requiring model retraining. Unlike static vector databases, which require periodic reprocessing, \name continuously integrates new information from multiple verified sources. Figure~\ref{fig:fig1}i illustrates this capability. Bizengri (zenocutuzumab-zbco) was approved by the FDA in December 2024, after the knowledge cutoff of \name's base model, Llama3.1-8B (December 2023). Instead of relying on outdated internal knowledge, \name calls the \textit{get\_indications} tool to query the openFDA API, retrieving the latest drug label information. This allows \name to correctly identify Bizengri's approved indications for non-small cell lung cancer and pancreatic adenocarcinoma. By integrating continuously updated sources, \name ensures access to the latest biomedical knowledge, eliminating reliance on static training data and mitigating knowledge obsolescence.

\subsection*{\toolbox: A universe of tools and machine learning models}

\toolbox is a suite of 211 biomedical tools that integrate with \name. It covers a wide range of categories (Figure~\ref{fig:fig1}c), including adverse events, risks, and safety; addiction and abuse; drug usage in specific populations; drug administration and handling; pharmacology; drug mechanisms and composition; ID and labeling tools; general clinical annotations; clinical laboratory information; patient and caregiver resources; pairwise disease, phenotype, target, and drug associations; biological annotation tools; publication information; search tools; and target characterization. Tools in \toolbox are built on APIs from trusted sources, including openFDA~\cite{kass2016openfda}, Open Targets~\cite{ochoa2023next}, and the Monarch Initiative~\cite{castellanos2024human}. \extfigref{fig:tool_dist} provides a detailed breakdown of \toolbox tools.

\xhdr{\toolgen agents generate a dataset of tool specifications used to create \toolbox.}
The \toolgen system constructs tools in \toolbox using a multi-agent approach that converts API documentation into structured tool specifications (\extfigref{fig:multi-agent-system}a). API documentation varies widely in format and content, making direct integration with \name challenging. \toolgen standardizes this process by organizing API functions into well-defined tools with clear, concise descriptions that \name can interpret. The system operates in four stages:
\begin{enumerate}[leftmargin=*,noitemsep,topsep=0pt] 
\item Capability summarization: The \textsc{Summarizer} agent extracts and condenses API documentation to identify the API’s core functionalities.
\item Tool generation: The \textsc{Tool Generator} agent translates these capabilities into structured tool specifications. Each tool specification includes a description for \name's function calls and a mapping rule that converts function calls into API requests. The tool description defines the tool’s name, purpose, input arguments, data types, and mandatory parameters (examples in Figure~\ref{fig:tool_description}b and \extfigref{fig:tool_description}).
\item Tool validation: The \textsc{Tool Checker} agent generates test cases with predefined queries and function calls to verify the tool’s functionality.
\item Human verification: Experts manually review and refine tools to ensure correctness, meaningful applications, and robustness to unexpected inputs.
\end{enumerate}
The \textsc{Summarizer}, \textsc{Tool Generator}, and \textsc{Tool Checker} agents operate by prompting the LLM with specialized instructions. \onlineref{sec:data_gen_method} provides additional details about the \toolgen system.

\subsection*{\trainset dataset of therapeutic tasks and reasoning traces}

We construct \trainset, a multi-step reasoning and function call training dataset (Figure~\ref{fig:fig2}d). \trainset consists of three datasets: a tooling dataset, a therapeutic question dataset, and a reasoning trace dataset, generated by three agent systems (\extfigref{fig:multi-agent-system}).
The tooling dataset contains augmented versions of 211 tools from \toolbox. Each tool description is rephrased to introduce variability, ensuring that \name learns tool usage rather than memorizing specific descriptions. The therapeutic question dataset includes 85,340 questions and functional instructions generated by the \questgen agent system to train \name's reasoning capabilities. The reasoning trace dataset comprises 85,340 detailed reasoning traces that contain 177,626 reasoning steps and 281,695 function calls, all generated by the \tracegen agent system.
Processing these three datasets (as detailed in \onlineref{sec:dataset_convert}) results in \trainset, which includes 378,027 instruction-tuning samples. The agent systems generate training data by sampling drugs and disease information from verified biomedical sources. Drug data is obtained from FDA drug labeling documents~\cite{kass2016openfda}, while disease information is sourced from PrimeKG~\cite{chandak2022building}. Drug-disease, phenotype, and target associations are compiled from Open Targets~\cite{ochoa2023next}.

\xhdr{\questgen agents generate a dataset of therapeutic questions.}
\questgen constructs therapeutic questions with treatment, disease, and drug-related information. Training \name requires a large dataset of questions that address various forms of therapeutic reasoning, including patient populations, drug side effects, and drug interactions. Manually generating these questions is infeasible. Instead, \questgen, a multi-agent system, generates meaningful questions from verified knowledge bases (\onlineref{sec:question_gen}, \extfigref{fig:multi-agent-system}b).
\questgen operates in three stages. First, the \textsc{Information Extractor} agent identifies and extracts key information from biomedical documents and data sources. Next, the \textsc{Question Generator} agent constructs questions using the extracted information and generates corresponding answers with detailed explanations that clarify how the answer addresses the question. Finally, \questgen evaluates each question based on knowledge grounding, solvability, and reasonableness. Only validated questions proceed to the \tracegen system for reasoning trace generation.

\xhdr{\tracegen agents generate a dataset of therapeutic reasoning traces.}
To generate valid reasoning traces that integrate feedback from real-world tools, we design \tracegen, a multi-agent system that constructs complex, step-wise reasoning traces (\extfigref{fig:multi-agent-system}c). \tracegen produces training data for each question, including a reasoning trace and the final answer. Generating reasoning traces presents several challenges:
(1) Complexity of questions: Many questions require multi-step reasoning and the analysis of multiple factors, making it difficult to generate a single direct answer. \tracegen must generate reasoning traces that effectively handle this complexity.
(2) Integration of external tools: Effective reasoning requires incorporating real-world tools rather than relying solely on the internal knowledge of LLMs. \tracegen must integrate tool outputs into reasoning traces while ensuring consistency across sources.
(3) Handling unpredictable tool outputs: External tools often produce unexpected results. \tracegen must manage failure cases, filter noisy outputs, and ensure that reasoning progresses toward a valid solution despite deviations in tool responses.
\tracegen addresses these challenges using a multi-agent system consisting of the \textsc{Helper} agent, the \textsc{Tool Provider} module, the \textsc{Solver} agents, and a reasoning trace evaluation step (\extfigref{fig:multi-agent-system}c).
\begin{itemize}[leftmargin=*,noitemsep,topsep=0pt] 
\item The \textsc{Helper} agent provides the \textsc{Solver} with step-by-step hints, guiding the reasoning process based on previous steps. It has access to correct answers and explanations, ensuring alignment with expected outcomes.
\item The \textsc{Tool Provider} module identifies relevant tools based on the question and recommendations from \toolrag, which iteratively improves tool selection accuracy by learning from previously generated data.
\item The \textsc{Solver} agent integrates information from the \textsc{Tool Provider}, \textsc{Helper}, and existing reasoning traces to iteratively generate reasoning steps and function calls until reaching a final answer.
\item The evaluation step verifies the correctness of the answer, function calls, and reasoning process while detecting hallucinations, arbitrary outputs, and repetitive reasoning patterns.
\end{itemize}
Details of \tracegen are provided in \onlineref{sec:trace_gen}.

\subsection*{\name outperforms larger LLMs in multi-step reasoning}
We construct the DrugPC (Drug Prescribing Card) benchmark to evaluate \name's performance in drug reasoning. DrugPC includes 3,168 questions spanning 11 tasks: drug overview, ingredients, warnings and safety, dependence and abuse, dosage and administration, use in specific populations, pharmacology, clinical information, nonclinical toxicology, patient-focused information, and storage and supply.
To mitigate data leakage from pretraining, we focus on drugs approved by the FDA in 2024, reducing the likelihood that LLMs have encountered them. 
We exclude drugs approved after 2023 from the training set and use drugs approved in 2024 for evaluation. We perform instruction tuning on LLMs, such as the Llama-3.1-8B-Instruct model with 8 billion parameters, using \trainset to develop \name's reasoning and tool-use capabilities. Training details are provided in \onlineref{sec:training_design}.
We evaluate models in two settings: multiple-choice, where the model selects the correct answer from given options, and open-ended, where the model generates responses without predefined choices. By default, \questgen generates questions with 4-5 options, verified by human experts. To create open-ended versions, we remove answer choices from the input. After generating a response, the model selects the correct option from the original choices based on its generated text.
Table~\ref{tab:example_question_type} provides examples of both formats. Further details on benchmark datasets and evaluation are in \onlineref{sec:benchmark_details}.

\name is built on the Llama-3.1-8B-Instruct model, which has 8 billion parameters and is fine-tuned for multi-step reasoning and function call execution. We compare \name to larger models, including Llama3.1-70B-Instruct (70 billion parameters) and GPT-4o (Figure~\ref{fig:fig1}d).
Despite its smaller size, \name consistently outperforms Llama3.1-70B-Instruct in both multiple-choice and open-ended tasks. In the multiple-choice setting, \name achieves 93.8\% accuracy, surpassing Llama3.1-70B-Instruct's 75.1\%. In the open-ended setting, \name maintains 92.1\% accuracy, while Llama3.1-70B-Instruct drops to 52.8\%.
Among baseline models, GPT-4o performs best, achieving 76.4\% in multiple-choice and 66.3\% in open-ended tasks. However, \name outperforms GPT-4o by 17.4\% in multiple-choice and 25.8\% in open-ended settings. By leveraging multi-step reasoning and executing function calls to \toolbox for verified information, \name surpasses larger models in accuracy and reliability.
The open-ended setting is more challenging than the multiple-choice format, as models cannot rely on answer choices. GPT-4o and Llama3.1-70B-Instruct show accuracy drops of 10.1\% and 22.3\%, respectively, when switching to open-ended tasks. In contrast, \name exhibits only a 1.7\% decline, highlighting its robustness in open-ended reasoning.

We evaluate performance across all 11 tasks in the DrugPC benchmark (Figure~\ref{fig:fig2}b,c). Although GPT-4o is the strongest baseline overall, it does not consistently outperform other models. For example, Llama3.1-70B-Instruct achieves higher accuracy than GPT-4o on the Warning and Safety task. In contrast, \name surpasses all baselines in all tasks, demonstrating its effectiveness in multitask drug reasoning. \name provides reasoning traces supported by verified function call results, allowing users to assess the reliability of the response. In contrast, LLM-generated outputs require manual verification, limiting trust without external validation.

\subsection*{\name outperforms tool-use LLMs in multi-step reasoning}

We compare \name with tool-use LLMs that support function calling~\cite{watt-tool-8B,liu2024toolace,berkeley-function-calling-leaderboard} (Figure~\ref{fig:fig1}e). Existing models focus on generating accurate function calls based on input questions and tool descriptions but lack the ability to handle complex problems requiring multi-step function calls, reasoning, and diverse tool integration. By incorporating multi-step reasoning and function call capabilities, \name provides key advantages over existing tool-use LLMs: (1) Expanded tool support: \name employs goal-oriented tool selection, enabling access to a large number of tools in \toolbox. In contrast, existing methods rely on including all tool descriptions in the context window, limiting the number of tools they can handle. Some tool-use LLMs~\cite{gorilla-openfunctions-v2} cannot support large-scale toolboxes like \toolbox. (2) Improved problem-solving: \name performs multi-round function calls to address complex problems. When a single function call does not provide sufficient information, \name reevaluates and selects alternative tools to refine its solution.

We compare \name against state-of-the-art tool-use LLMs, including ToolACE-8B~\cite{liu2024toolace} and WattTool-8B~\cite{watt-tool-8B}, both fine-tuned on the same Llama-3.1-8B-Instruct model as \name. To ensure a fair comparison, we provide all models with full access to \toolbox and enable multi-step reasoning. Since existing tool-use LLMs do not natively support multi-step reasoning but allow multi-round interactions, we simulate multi-step reasoning by feeding tool results back as user messages, allowing the LLM to continue function calls until reaching a final answer. Additionally, because most tool-use LLMs struggle with switching between function calls and answer generation, we introduce a special \textsc{GiveAnswer} tool. This tool requires the model to invoke it with the final answer once problem-solving is complete, ensuring a structured response process.

\name achieves significantly higher accuracy than existing tool-use LLMs. In the multiple-choice setting, \name outperforms ToolACE by 62.5\% and WattTool by 59.1\%. In the open-ended setting, \name surpasses ToolACE by 59.4\% and WattTool by 55.0\%. This performance gap arises from key limitations in existing tool-use LLMs: (1) Limited tool selection: These models struggle to handle many tools in a single context window and often fail to select the correct tool from hundreds available in \toolbox. (2) Single-round function calls: They fill in function arguments based only on the input question, without making additional calls to retrieve missing information. (3) Ineffective multi-step reasoning: Lacking adaptive reasoning, they often repeat initial function calls instead of refining their approach based on previous results, leading to failures when reaching the maximum reasoning round limit.

We quantify these failures by tracking invalid answers—cases where the model cannot produce a valid response. WattTool-8B fails on 58.9\% of multiple-choice and 56.6\% of open-ended questions. ToolACE-8B fails on 63.1\% and 60.7\% of multiple-choice and open-ended questions, respectively. In contrast, \name employs multi-step reasoning, iterative function calls, and goal-oriented tool selection, allowing it to fully use \toolbox in therapeutic reasoning.

\subsection*{\name generalizes across drug name variants and descriptions}

We evaluate \name's ability to generalize across different drug representations. LLM-based models are sensitive to variations in how drugs are referenced~\cite{gallifant2024language}, such as brand versus generic names. To test generalization, we construct three modified versions of the DrugPC benchmark: BrandPC, GenericPC, and DescriptionPC.
BrandPC and GenericPC systematically replace drug names in DrugPC with their brand or generic equivalents. Questions that do not reference drug names remain unchanged, while those requiring conversion between brand and generic names are modified accordingly. Both datasets maintain the same number of samples as DrugPC. Sample questions are shown in Figure~\ref{fig:fig3}a.

DescriptionPC replaces drug names with detailed descriptions to assess generalization without explicit drug names, including indications, mechanisms of action, contraindications, and interactions. We removed DrugPC questions that became unanswerable after this transformation, resulting in 626 questions.
Since multiple drugs may share similar descriptions, DescriptionPC introduces a two-step evaluation: (1) drug identification and (2) answer correctness (Figure~\ref{fig:fig3}b). In the first step, the model identifies the drug based on its description. The ground truth includes all drugs that could match the given description. In the second step, the model selects the correct answer to a multiple-choice question using its predicted drug name. If drug identification is incorrect, the answer is automatically marked incorrect, ensuring that predictions rely on accurate drug recognition.

\name achieves 93.6\% accuracy on BrandPC and 93.7\% on GenericPC, outperforming both pure LLMs and tool-use LLMs on both benchmarks (Figure~\ref{fig:fig3}a).
Among pure LLMs, Llama3.1-70B-Instruct performs best on BrandPC (73.0\%), while GPT-4o leads on GenericPC (77.3\%). \name surpasses these top reference models by 20.6\% and 16.4\%, respectively.
Among tool-use LLMs, WattTool-8B achieves the highest accuracy, with 40.2\% on BrandPC and 31.5\% on GenericPC. \name outperforms these baselines by 53.4\% and 62.2\%, respectively.
\name also exhibits lower performance variance across the original, BrandPC, and GenericPC datasets, with a variance of 0.00667. In contrast, GPT-4o has a variance of 9.96, Llama3.1-70B-Instruct 2.42, WattTool-8B 13.07, and ToolACE-8B 1.05. These results demonstrate \name's superior robustness and generalization across different drug name representations.

On the DescriptionPC benchmark (Figure~\ref{fig:fig3}b), when evaluating only answer correctness (without considering whether the model identifies the correct drug) \name achieves 90.4\%, surpassing GPT-4o (85.9\%) and Llama3.1-70B-Instruct (85.3\%). However, models may be able to ``guess" the answer to certain questions in DescriptionPC without first identifying the class of drugs being referenced, which limits model trustworthiness. Specifically, when requiring both correct drug identification and answer selection, accuracy drops significantly for Llama3.1-70B-Instruct to 20.1\%, indicating unreliable drug grounding. In contrast, \name maintains the highest performance at 56.5\%, outperforming GPT-4o by 8.3\%.
For drug name identification alone, \name achieves the highest accuracy at 60.1\%, compared to GPT-4o's 55.8\% and Llama3.1-70B-Instruct's 23.6\%. These results highlight \name's stronger ability to reason over drugs and base decisions on correct information.

\subsection*{\name for precision treatment recommendation}

We evaluate \name's ability to provide personalized treatment recommendations using the TreatmentPC benchmark, which consists of 456 questions focused on specialized treatment scenarios. While multiple drugs may treat a single disease, patient-specific factors (such as pregnancy or comorbidities) require tailored drug selection and dosage adjustments. TreatmentPC assesses these cases by formulating questions that account for varying drug application conditions. We select drugs approved by the FDA in 2024, identify their indicated diseases, and analyze treatment options by comparing drug attributes. For example, among all available treatments, only one drug may be suitable for pregnant patients. This analysis is based on FDA documentation, including indications, usage in specific populations, safety warnings, precautions, and contraindications.

Using these drug-specific properties, we generate multiple-choice questions with 4-5 options, ensuring only one correct choice based on the patient’s condition. Questions also include scenarios where drug interactions must be considered, requiring the model to account for contraindications. We evaluate models in both multiple-choice and open-ended settings. In the multiple-choice format, the model selects the most appropriate drug from the given options. In the open-ended format, the model generates a treatment recommendation and later selects the correct answer from its own response. TreatmentPC measures \name's ability to analyze patient conditions and recommend appropriate treatments. Further details on the benchmark dataset and evaluation methodology are in \onlineref{sec:benchmark_details}.

\xhdr{\name outperforms LLMs and tool-use LLMs in TreatmentPC.}
\name achieves significantly higher accuracy than its fine-tuning base model, Llama-3.1-8B-Instruct (Figure~\ref{fig:fig4}a). In the multiple-choice setting, \name reaches 86.8\% accuracy, surpassing Llama-3.1-8B-Instruct's 56.1\%. In the open-ended setting, \name attains 75.0\%, outperforming Llama-3.1-8B-Instruct's 33.11\%. Compared to larger LLMs, \name maintains superior performance. In the multiple-choice setting, it outperforms GPT-4o by 12.7\% and Llama-3.1-70B-Instruct by 16.4\%. The gap widens in the open-ended setting, where \name exceeds GPT-4o by 13.6\% and Llama-3.1-70B-Instruct by 25.4\%. Even in the open-ended setting, \name (75.0\%) surpasses GPT-4o's multiple-choice accuracy (74.1\%), despite the latter benefiting from predefined answer choices.

\name also outperforms tool-use LLMs (Figure~\ref{fig:fig4}b). ToolACE-8B and WattTool-8B, fine-tuned on the same Llama-3.1-8B-Instruct model and given full access to \toolbox, perform significantly worse. In the multiple-choice setting, WattTool-8B achieves only 18.2\%, while \name reaches 86.8\%. In the open-ended setting, ToolACE-8B scores 13.4\%, compared to \name's 75.0\%. As observed in DrugPC, \name's advantage stems from its multi-step reasoning capabilities. It integrates information from multiple sources, executes iterative function calls, refines queries when initial tool calls return empty results, and dynamically adjusts its approach. These strengths enable \name to solve complex treatment recommendation tasks more effectively than existing tool-use LLMs.

\xhdr{\name outperforms reasoning LLMs, including DeepSeek-R1.} 
Recent reasoning LLMs, such as DeepSeek-R1~\cite{guo2025deepseek} and GPT-o1~\cite{OpenAI2024}, are designed for long chain-of-thought reasoning and test-time scaling. Since TreatmentPC requires multi-step reasoning over patient conditions and drug effects, we compare \name with DeepSeek-R1 models (Figure~\ref{fig:fig4}c).
To enable multi-step reasoning in DeepSeek-R1, we explicitly prompt it to generate reasoning steps using special tokens \textless think\textgreater{} and \textless\textbackslash think\textgreater{}. Despite DeepSeek-R1's full model having 671 billion parameters, \name outperforms it by 10.3\% in the multiple-choice setting (86.8\% vs. 76.5\%) and by 7.5\% in the open-ended setting.
\extfigref{fig:comparison_deepseek} shows the comparison between Deepseek-R1 and TxAgent.
Deepseek-R1 relies on internal knowledge for reasoning, risking hallucinations and misjudgments.
In contrast, TxAgent bases its reasoning on trusted sources, such as FDA drug labels, minimizing the risk of hallucinations and ensuring more reliable conclusions.

\name also surpasses distilled variants, including DeepSeek-R1-Llama-8B/70B, which are trained on Llama-3.1-8B and Llama-3.1-70B. Against DeepSeek-R1-Llama-8B, which shares the same base model as \name, \name achieves accuracy gains of 36.1\% in multiple-choice and 34.9\% in open-ended tasks.
Unlike reasoning LLMs that rely solely on internal knowledge, \name integrates multi-step reasoning with verified external information from \toolbox, making it more effective for specialized treatment recommendation tasks.

\subsection*{Examples of \name reasoning traces for specialized treatments}

We present detailed \name reasoning traces for four personalized treatment questions, evaluating its ability to incorporate drug mechanisms, drug-drug interactions, comorbidities, and clinical guidelines for specific patient groups, including elderly and pediatric patients.

\xhdr{(1) Treatment selection based on drug mechanism and pediatric use.}
Figure~\ref{fig:fig4}d presents a case of a pediatric male patient with Duchenne muscular dystrophy (DMD) seeking treatment. The patient does not want steroid-based therapies due to side effects, including weight gain and mood changes~\cite{gloss2016practice}, and is ineligible for exon-skipping antisense oligonucleotides, which are effective only for specific genetic mutations~\cite{matsuo2021antisense}. \name must identify an appropriate non-steroidal, non-exon-skipping treatment.
\name first calls \toolrag to find tools that identify drugs based on indications. It selects \textit{get\_drug\_names\_by\_indication} and retrieves ten DMD drugs. Analyzing the results, \name determines that Duvyzat is the only drug meeting the patient's criteria.
To assess pediatric suitability, \name calls \textit{get\_drug\_name\_by\_pediatric\_use}, but this tool does not return relevant information. \name then queries \toolrag again for tools related to pediatric guidelines and selects \textit{get\_pediatric\_use\_by\_drug\_name}, which confirms that Duvyzat is safe for children over six years old. Based on this reasoning, \name confidently recommends Duvyzat for this patient.
This case study highlights \name's ability to distinguish drugs by mechanism despite similar indications and to integrate mechanistic considerations with personalized factors such as pediatric safety.

\xhdr{(2) Treatment selection considering drug-drug interactions.}
Figure~\ref{fig:fig4}e examines a treatment decision that involves drug-drug interactions. The patient is currently taking Prozac (fluoxetine hydrochloride) for Major Depressive Disorder and is considering adding Xolremdi (mavorixafor) for WHIM syndrome. \name assesses whether these medications can be taken together.
\name first queries \toolrag for tools related to drug indications and contraindications. It simultaneously calls \textit{get\_indications} and \textit{get\_contraindications} for Xolremdi. The retrieved data confirm that Xolremdi is indicated for WHIM syndrome but is contraindicated with drugs dependent on CYP2D6 for clearance. Xolremdi inhibits CYP2D6, reducing its enzyme activity and prolonging the presence of drugs metabolized by CYP2D6 in the body.

To determine whether this contraindication applies to Prozac, \name calls \textit{get\_drug\_inter\-actions}, which reveals that Prozac is both a substrate and an inhibitor of CYP2D6. This presents two potential drug-drug interactions: (1) Direct contraindication: Prozac is metabolized by CYP2D6, and since Xolremdi inhibits this enzyme, Prozac exposure would increase, potentially leading to adverse effects. (2) Compounded inhibition: Both Prozac and Xolremdi reduce CYP2D6 activity. Their combined effect could further affect the metabolism of CYP2D6, increasing exposure to other drugs metabolized with CYP2D6.
Based on these interactions, \name concludes that taking Prozac and Xolremdi together is not suitable for the patient. This case highlights \name's ability to analyze drug-drug interactions through multi-step reasoning and detailed biological insights retrieved from \toolbox.

\xhdr{(3) Treatment selection considering geriatric use.} 
Figure~\ref{fig:fig4}f examines \name's ability to consider geriatric-specific treatment adjustments. A 70-year-old patient with schizophrenia seeks the maximum recommended dosage for Cobenfy (xanomeline and trospium chloride), a recently approved drug. Since the dosage can be adjusted based on patient response, \name must determine the appropriate upper limit and provide justification.
\name first calls \toolrag to retrieve relevant dosage- and age-related tools from \toolbox. It then selects and executes \textit{get\_dosage\_and\_storage\_info} and \textit{get\_geriatric\_use\_info}. The dosage tool confirms that the maximum recommended dose for elderly patients is 100 mg/20 mg twice daily, lower than the 125 mg/30 mg twice daily recommended for younger patients. The geriatric use tool explains that this adjustment is due to an increased risk of urinary retention in elderly patients.
\name synthesizes these findings and provides a final answer with supporting evidence. This case study highlights \name's ability to conduct parallel reasoning threads—both identifying and explaining the maximum dosage—by executing multiple tool calls simultaneously. It also demonstrates how \name integrates verified external information to deliver patient-specific, evidence-based treatment recommendations.

\xhdr{(4) Treatment selection considering comorbidities.} 
Figure~\ref{fig:fig4}g demonstrates \name's ability to incorporate comorbidities into treatment recommendations. The patient has two cardiac conditions: second-degree AV block, which disrupts electrical signaling in the heart, and hypertension. \name is tasked with identifying an appropriate hypertension treatment while considering the AV block.
\name first retrieves indication-related tools using \toolrag. It calls \textit{get\_drug\_names\_by\_indication} to identify ten potential hypertension treatments. Next, it filters these candidates based on contraindications for AV block. Using \textit{get\_drug\_name\_by\_contraindication} with the argument ``AV block,'' \name searches FDA-approved drug labels for contraindications. The results show that two of the retrieved hypertension drugs are contraindicated for patients with second-degree AV block.
\name then summarizes the mechanisms of the non-contraindicated drugs and provides them as the final answer. This case study highlights \name's ability to integrate comorbidity considerations into treatment recommendations and efficiently search and filter drug candidates using FDA drug labels.

\subsection*{Impact of tools in \toolbox on \name's performance}

We evaluate two factors: the reliability of tools compared to language model-based alternatives and the effect of expanding \toolbox on agent's performance.

\xhdr{\toolbox tools provide more accurate information than LLMs.}
\toolbox improves \name's reasoning accuracy by integrating verified knowledge sources through specialized tools. We compare its effectiveness against an LLM-only approach, where the model mimics tool functionality by receiving structured prompts that describe each tool’s capabilities and arguments (Figure~\ref{fig:fig3}c, \onlineref{sec:llm_as_tool}). GPT-4o and Llama 3.1-Instruct-8B serve as the backend LLMs in this analysis, with all other settings unchanged.
Replacing real tools in \toolbox with LLM-based tools significantly reduces accuracy. On DrugPC, using Llama3.1-8B-Instruct as tools lowers accuracy from 93.8\% to 68.7\% (-25.1\%), while using GPT-4o results in 72.7\% (-21.1\%). Although GPT-4o performs better, both models remain inferior to \toolbox, demonstrating the limitations of LLM-only approaches in retrieving precise biomedical information.
We observe a similar pattern on TreatmentPC. GPT-4o and Llama3.1-8B-Instruct achieve 67.11\% and 74.78\% accuracy, respectively, compared to 86.84\% with real tools in \toolbox. While advanced LLMs improve factual consistency, they still underperform compared to real-world tools. \toolbox ensures verifiable results, allowing users to validate \name's reasoning trace and final outputs.

\xhdr{Scaling \toolbox improves performance.}
We evaluate the effectiveness and scalability of \toolbox by measuring how performance changes as more tools are added. We construct four subsets containing 10\%, 20\%, 50\%, and 75\% of \toolbox, ensuring each larger subset includes all tools from the smaller ones. This approach allows us to assess the incremental impact of adding tools while maintaining continuity across evaluations. Using \name equipped with each subset and the full \toolbox, we measure accuracy on the DrugPC and TreatmentPC benchmarks (Figure~\ref{fig:fig3}d). Accuracy on DrugPC increases from 78.4\% with 10\% of the tools to 93.8\% with the full selection. A similar trend is observed on TreatmentPC, where accuracy rises from 71.7\% to 86.8\%. These results demonstrate that expanding \toolbox consistently improves \name's ability to handle complex, specialized treatment tasks.

\subsection*{The critical role of reasoning in \name}

This section evaluates the role of reasoning in \name through three experiments. First, we assess the impact of thought generation by removing this process. Second, we examine how the number of reasoning steps in training data affects performance by limiting the maximum reasoning traces. Finally, we evaluate the influence of reasoning during inference by imposing a step limit, forcing \name to generate a final answer after a predefined number of steps.

\xhdr{Explicit thought generation drives reasoning in \name.}
We evaluate the impact of thought generation by comparing \name with and without this process on the DrugPC and TreatmentPC benchmarks, using accuracy as the metric (Figure~\ref{fig:fig3}e). Unlike existing tool-use LLMs that generate only function calls, \name produces both reasoning thoughts and function calls at each step. To assess the importance of thought generation, we modify \name to generate only function calls without intermediate reasoning. At the final step, it directly outputs the answer instead of reasoning through function calls. We implement this by removing the thought process from the \trainset dataset (\onlineref{sec:no_thought_inference}). Eliminating thought generation reduces accuracy on DrugPC from 93.8\% to 71.5\% (-22.3\%) and on TreatmentPC from 86.4\% to 64.9\% (-21.5\%). This decline demonstrates the critical role of explicit reasoning in \name and its advantage over tool-use LLMs that rely solely on function calls.

\xhdr{Long multi-step training traces improve performance on complex tasks.}
We evaluate how the number of reasoning steps in \name's training data affects its performance on the DrugPC and TreatmentPC benchmarks, using accuracy as the metric (Figure~\ref{fig:fig3}f). \name acquires multi-step reasoning through fine-tuning on the \trainset dataset. To assess the impact of reasoning depth, we filter training data to retain samples with at most 1, 3, or 5 reasoning steps, or all available steps (\onlineref{sec:dataset_convert}). During inference, \name remains unrestricted in the number of reasoning steps it can take. Reducing reasoning steps in training significantly decreases performance. A model trained with only 1 reasoning step sees accuracy drop from 86.8\% to 66.9\% on TreatmentPC and from 93.8\% to 71.6\% on DrugPC. The decline is more pronounced on TreatmentPC, indicating that complex treatment decisions require stronger multi-step reasoning. These results demonstrate that deeper reasoning traces during training improve \name’s ability to handle complex therapeutic tasks.

\xhdr{Longer inference traces improve performance.}
We assess the impact of reasoning depth during inference by imposing a step limit on \name, using the TreatmentPC benchmark and accuracy as the evaluation metric (Figure~\ref{fig:fig3}g). \name is trained on the full \trainset dataset but is restricted to a maximum number of reasoning steps at inference. As described in Algorithm~\ref{alg:txagent_inference}, instead of allowing \name to autonomously determine when to generate the special token \texttt{[FinalAnswer]}, we enforce this token when \name reaches the step limit, instructing it to produce the final answer based on the accumulated reasoning trace. For reasoning traces shorter than the limit, the inference process remains unchanged. Accuracy improves as the reasoning step limit increases. When restricted to a single step—equivalent to conventional LLMs that generate direct answers—\name achieves 73.5\% accuracy, 13.3\% lower than its unrestricted multi-step reasoning configuration. Performance continues to improve with additional steps, showing notable gains up to five steps, after which improvements plateau. The diminishing returns beyond five steps suggest that most essential reasoning occurs within this range, though maintaining full reasoning capacity remains optimal.

As a reference, we present the average number of reasoning steps and tool calls for \name in~\extfigref{fig:avg_infer_step}. The TreatmentPC benchmark requires more reasoning steps than the DrugPC benchmarks, suggesting that precise treatment recommendations require more reasoning steps before reaching a conclusion.
Similarly, the TreatmentPC benchmark involves a greater number of tool calls compared to DrugPC. When comparing multiple-choice and open-ended settings, DrugPC shows no significant difference in reasoning steps or tool calls. However, in the open-ended setting, TreatmentPC requires a significantly larger number of reasoning steps and tool calls compared to the multiple-choice setting.

\section*{Discussion}


\name is an AI agent that applies multi-step reasoning and tool usage to solve therapeutic problems, including drug prescriptions and disease treatment recommendations, while considering patient-specific factors. Unlike conventional models that produce probability scores without explanations, \name generates a reasoning trail along with its answer, making its decision-making process transparent and interpretable. \name integrates external tools from \toolbox to retrieve real-time biomedical knowledge, overcoming the limitations of LLMs that rely solely on static training data. This enables \name to recommend newly approved drugs, assess indications, and provide evidence-based prescriptions. By grounding the responses in verified sources, \name allows users to trace each decision step in a transparent manner.

Treatment decisions must account for patient-specific factors, including age, comorbidities, pregnancy status, disease severity, and immune function. Existing models predict disease-drug links but fail to consider these variations. \name addresses this limitation through dynamic, multi-step reasoning. It identifies the disease based on phenotypes, retrieves potential treatments by considering associated phenotypes and biological targets, and evaluates drug suitability based on patient characteristics. Rather than following a fixed sequence, \name adapts its reasoning through iterative function calls to biomedical tools, ensuring decisions are grounded in verified sources such as FDA drug labels. For example, \name determines that Xolremdi, a treatment for WHIM syndrome, should not be used with Prozac, a CYP2D6 inhibitor, because it alters Xolremdi’s metabolism. By integrating patient-specific constraints into its reasoning process, \name ensures clinically relevant and personalized treatment recommendations.

\name's limitations highlight areas for future research. It relies on tool calls for external information, but gaps in \toolbox restrict access to specific data types, limiting its ability to address a broader range of questions.
Uncertainty quantification in \name's internal knowledge remains a challenge. The current approach grounds reasoning through external tools, improving verifiability. However, integrating internal knowledge with tool feedback could enhance flexibility for exploratory tasks.
\name processes only natural language inputs and does not yet support other modalities such as pathology images, EHR data, or web-based lab results. Expanding multi-modal support would enable \name to handle more complex cases and specialized clinical analyses.

\name is an AI agent for therapeutic reasoning that leverages a universe of tools to generate transparent reasoning traces grounded in multi-source medical evidence and continuously updated medical knowledge. It integrates verified information from FDA drug labels, Open Targets, and other trusted sources to produce evidence-based therapeutic recommendations. Future advances in integrating clinical modalities and extended memory for patient histories could allow \name to analyze multi-modal clinical data~\cite{su2025multimodal}. \name establishes a new framework for precision therapeutics by advancing personalized therapy selection and supporting regulatory-compliant clinical decision-making. 

\clearpage

\xhdr{Data and code availability.}
The project page is available at \url{https://zitniklab.hms.harvard.edu/TxAgent}.
The code and demo of \name are available at \url{https://github.com/mims-harvard/TxAgent}. 
The code of \toolbox is available at \url{https://github.com/mims-harvard/ToolUniverse}.
The pre-trained models are available at \url{https://huggingface.co/collections/mims-harvard/txagent-67c8e54a9d03a429bb0c622c}.

\xhdr{Acknowledgements.} 
We gratefully acknowledge the support of NIH R01-HD108794, NSF CAREER 2339524, US DoD FA8702-15-D-0001, Harvard Data Science Initiative, Amazon Faculty Research, Google Research Scholar Program, AstraZeneca Research, Roche Alliance with Distinguished Scientists, Sanofi iDEA-iTECH, Pfizer Research, Gates Foundation (INV-079038), Chan Zuckerberg Initiative, John and Virginia Kaneb Fellowship at Harvard Medical School, Biswas Computational Biology Initiative in partnership with the Milken Institute, Harvard Medical School Dean's Innovation Fund for the Use of Artificial Intelligence, and Kempner Institute for the Study of Natural and Artificial Intelligence at Harvard University.  Any opinions, findings, conclusions or recommendations expressed in this material are those of the authors and do not necessarily reflect the views of the funders.
We thank Owen Queen and Thomas Hartvigsen for their helpful discussion and feedback on our project.

DISTRIBUTION STATEMENT A. Approved for public release. Distribution is unlimited. This material is based upon work supported by the Under Secretary of Defense for Research and Engineering under Air Force Contract No.~FA8702-15-D-0001. Any opinions, findings, conclusions or recommendations expressed in this material are those of the author(s) and do not necessarily reflect the views of the Under Secretary of Defense for Research and Engineering.

\xhdr{Competing interests.} 
The authors declare no competing interests.

\clearpage


\newgeometry{left=0.7in,right=0.7in}
\captionsetup{margin=0.1in}
\spacing{1}
\pagestyle{empty}

\begin{figure}[ht]
\centering
\includegraphics[width=0.9\textwidth]{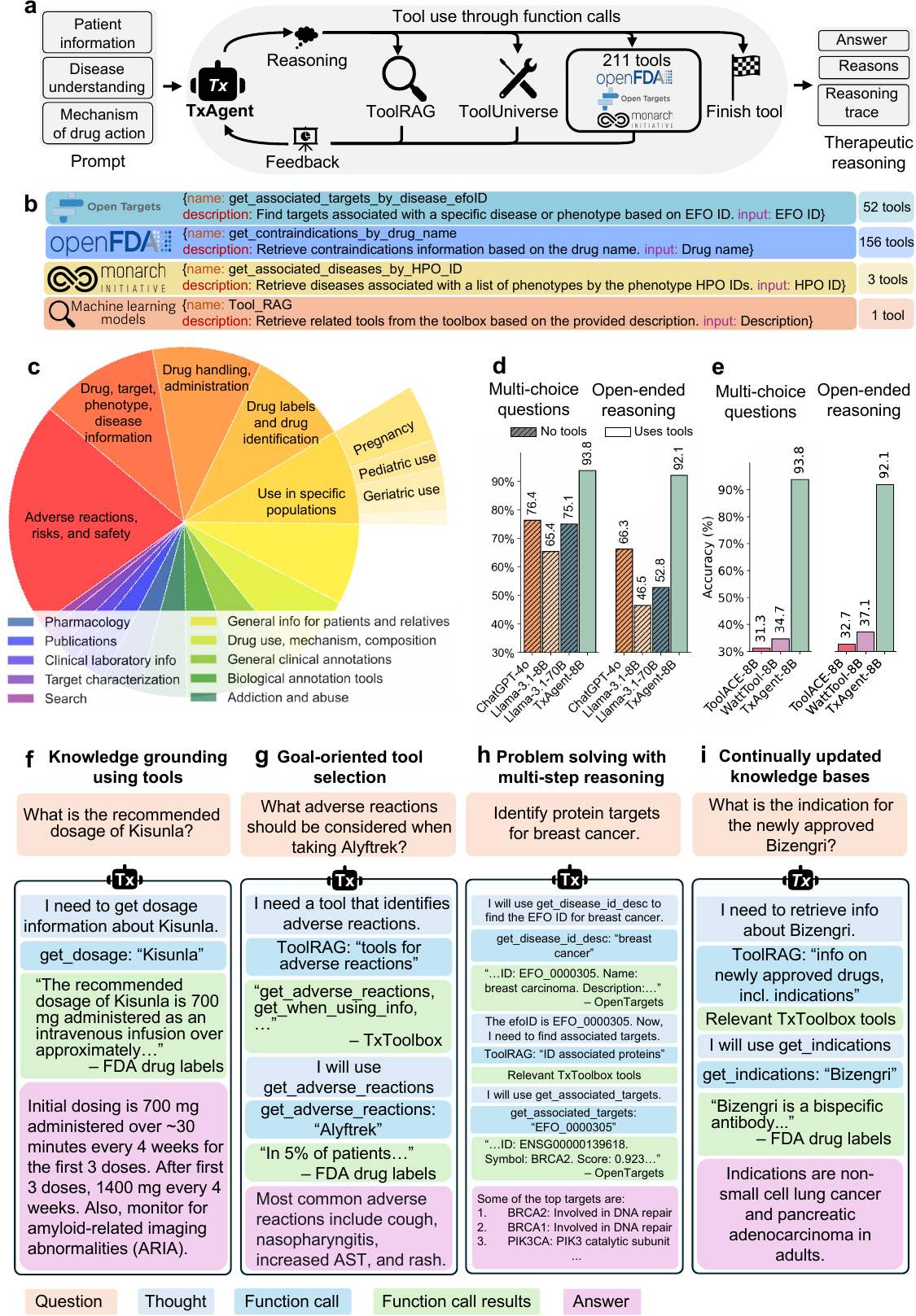}
\caption{}
\label{fig:fig1}
\end{figure}
\clearpage
\xhdr{Figure~\ref{fig:fig1}}:
\textbf{a})  
\name processes specialized therapy-related questions, generating detailed step-by-step reasoning and conducting parallel function calls across a vast array of biomedical tools and specialized tools. It delivers solutions supported by clear, rational, and verified reasoning traces.
\textbf{b}) Examples of tools in \toolbox and the machine learning tool. \toolbox consolidates 211 tools linked to trusted sources, including all US FDA-approved drugs since 1939 and validated clinical insights from Open Targets and Monarch Initiative.
The machine learning tool, e.g., \toolrag, is based on a machine learning model instead of APIs.
\textbf{c}) \toolbox includes 211 biomedical tools that address various aspects of drugs and diseases.
It covers the following categories: 
adverse events, risks, safety; addiction and abuse; drug usage in patient populations; drug administration and handling; pharmacology; drug use, mechanism, composition; ID and labeling tools; general clinical annotations; clinical laboratory info; general info for patients and relatives; disease, phenotype, target, drug links; biological annotation tools; publications; search; target characterization.
\textbf{d}) \name demonstrates superior performance compared to LLMs with a larger number of parameters, such as GPT-4o, excelling in both open-ended and multiple-choice questions.
\textbf{e}) \name demonstrates superior performance compared to tool-use LLMs that also have full access to \toolbox,  excelling in both open-ended and multiple-choice questions.
\textbf{f-i}) Capabilities of \name: knowledge grounding using tool calls, goal-oriented tool selection,
problem solving with multi-step reasoning, and leverage constantly updated knowledge base.
\textbf{f}) Knowledge grounding using tool calls, where \name utilizes tools to obtain verified knowledge and provides outputs based on it.
\textbf{g}) Goal-oriented tool selection, where \name proactively requests tools from \toolbox using the \toolrag model and selects and applies the most suitable tool from the available candidates.
\textbf{h}) Problem solving with multi-step reasoning, where \name manages complex tasks or unexpected responses from tools through multiple iterations of thought and function calls.
\textbf{i}) Leveraging constantly updated knowledge bases, where \name accesses continuously updated databases via tools to handle problems that go beyond the \name's intrinsic knowledge.
\clearpage

\begin{figure}[ht]
\centering
\includegraphics[width=0.9\textwidth]{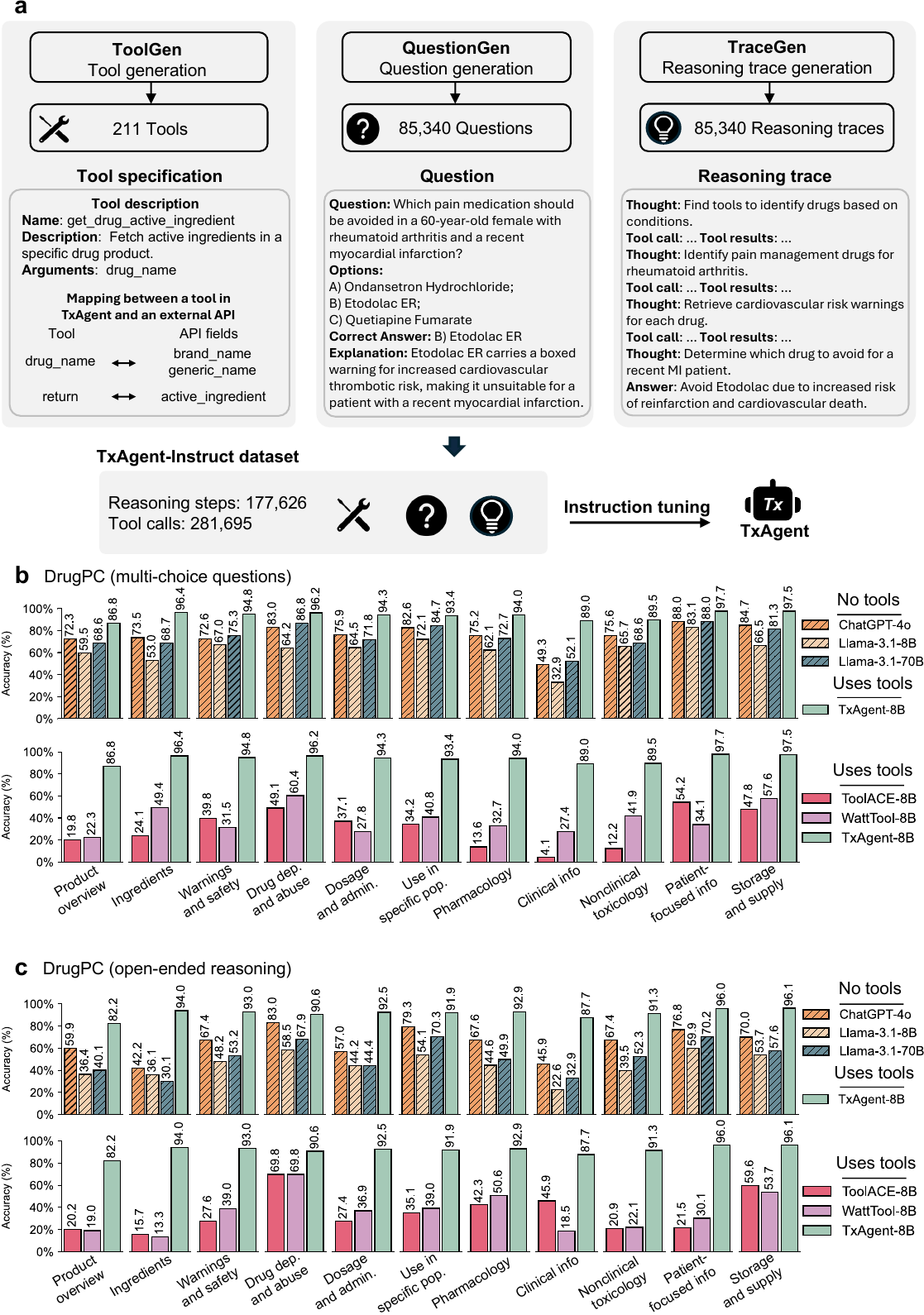}
\caption{
}
\label{fig:fig2}
\end{figure}
\clearpage
\xhdr{Figure~\ref{fig:fig2}}:
\textbf{a})
\trainset dataset is a diverse synthetic multi-step reasoning and massive function call training dataset anchored in biomedical knowledge.
To generate \trainset, we construct three datasets—a tooling dataset, a comprehensive therapeutic question dataset, and a reasoning trace dataset—using the auxiliary agent systems.
The tooling dataset consists of augmented versions of 211 tools from \toolbox.
The comprehensive therapeutic question dataset includes 85,340 therapeutic questions and functional instructions designed to train \name's abilities. These are generated by the \questgen agent system.
The reasoning trace dataset comprises 85,340 detailed reasoning traces for answering therapeutic questions. These traces collectively encompass 177,626 reasoning steps and 281,695 function calls, all generated by the \tracegen agent system.
By processing the data from these three datasets, we construct \trainset, which comprises 378,027 instruction-tuning data samples.
\textbf{b}) \name outperforms larger open-source LLMs and GPT-4 across 11 tasks from the DrugPC dataset, excelling in both open-ended and multiple-choice questions. These tasks cover various drug-related topics, including drug overview, ingredients, warnings and safety, dependence and abuse, dosage and administration, use in specific populations, pharmacology, clinical information, nonclinical toxicology, patient-focused information, and storage and supply.
\textbf{c}) Across the 11 tasks of the DrugPC dataset, \name demonstrates superior performance compared to existing tool-use LLMs.

\clearpage

\begin{figure}[ht]
\centering
\includegraphics[width=0.9\textwidth]{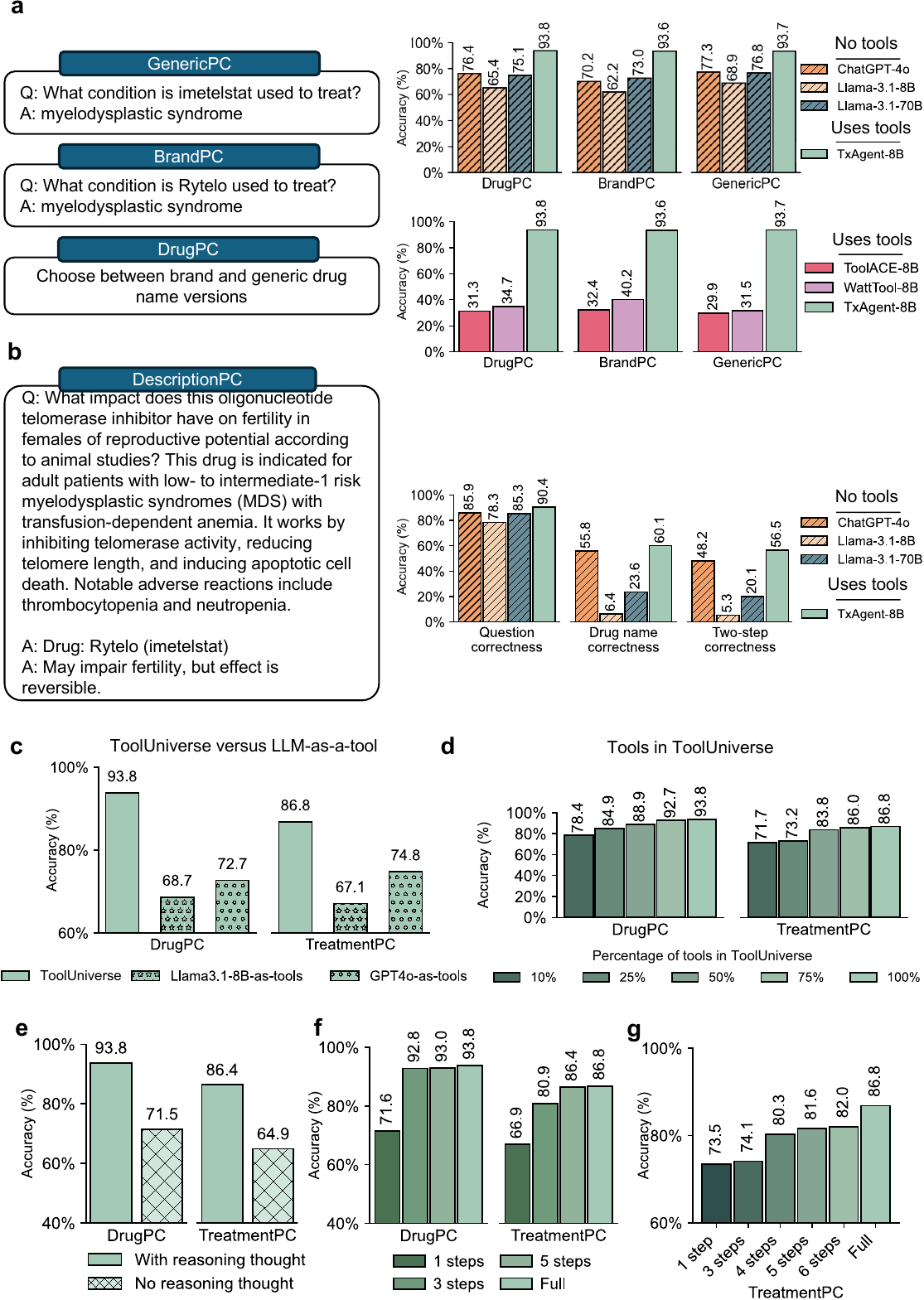}
\caption{
}
\label{fig:fig3}
\end{figure}
\clearpage
\xhdr{Figure~\ref{fig:fig3}}:
\textbf{a}) \name surpasses both native and tool-use LLMs on the DrugPC benchmark, as well as its Brand and Generic variants, where drug names are replaced with their brand and generic counterparts. Additionally, \name demonstrates minimal variance when handling drug names with different representations.
\textbf{b}) \name surpasses LLM in a two-step evaluation on the DescriptionPC benchmark, where drug names are replaced with their descriptions, including indications, mechanisms of action, contraindications, and interactions. In this evaluation, the first step involves identifying the correct drug name based on its description, followed by answering the question using the correctly identified drug name.
\textbf{c}) Comparison of real-world tools from \toolbox versus relying on an LLM’s internal knowledge as a substitute for external tools on DrugPC and TreatmentPC benchmarks. When paired with \name, \toolbox tools provide more accurate information than using LLMs ike GPT-4o as tools.
\textbf{d}) The impact of increasing the number of tools in \toolbox on the DrugPC and TreatmentPC benchmarks. As more tools are incorporated into \toolbox, the results consistently demonstrate steady and significant performance improvements.
\textbf{e}) Explicit thought generation is fundamental to reasoning in \name. We evaluate \name with and without thought generation on the DrugPC and TreatmentPC benchmarks. The absence of thought generation results in a significant performance decline, underscoring its essential role in \name’s reasoning process.
\textbf{f}) Long multi-step traces in training data enhance \name’s ability to handle complex tasks. We examine how the number of reasoning steps in \name’s training data affects its performance on the DrugPC and TreatmentPC benchmarks. As the number of reasoning steps decreases, performance gradually declines, suggesting that more complex tasks demand a stronger multi-step reasoning capability from \name.
\textbf{g}) Longer inference traces enhance model performance. To assess the impact of reasoning during inference, we impose a step limit on \name and evaluate its performance on the TreatmentPC benchmark. Results show a clear upward trend in accuracy as the number of reasoning steps increases, highlighting the importance of extended reasoning in \name’s inference process.
\clearpage

\begin{figure}[ht]
\centering
\includegraphics[width=0.8\textwidth]{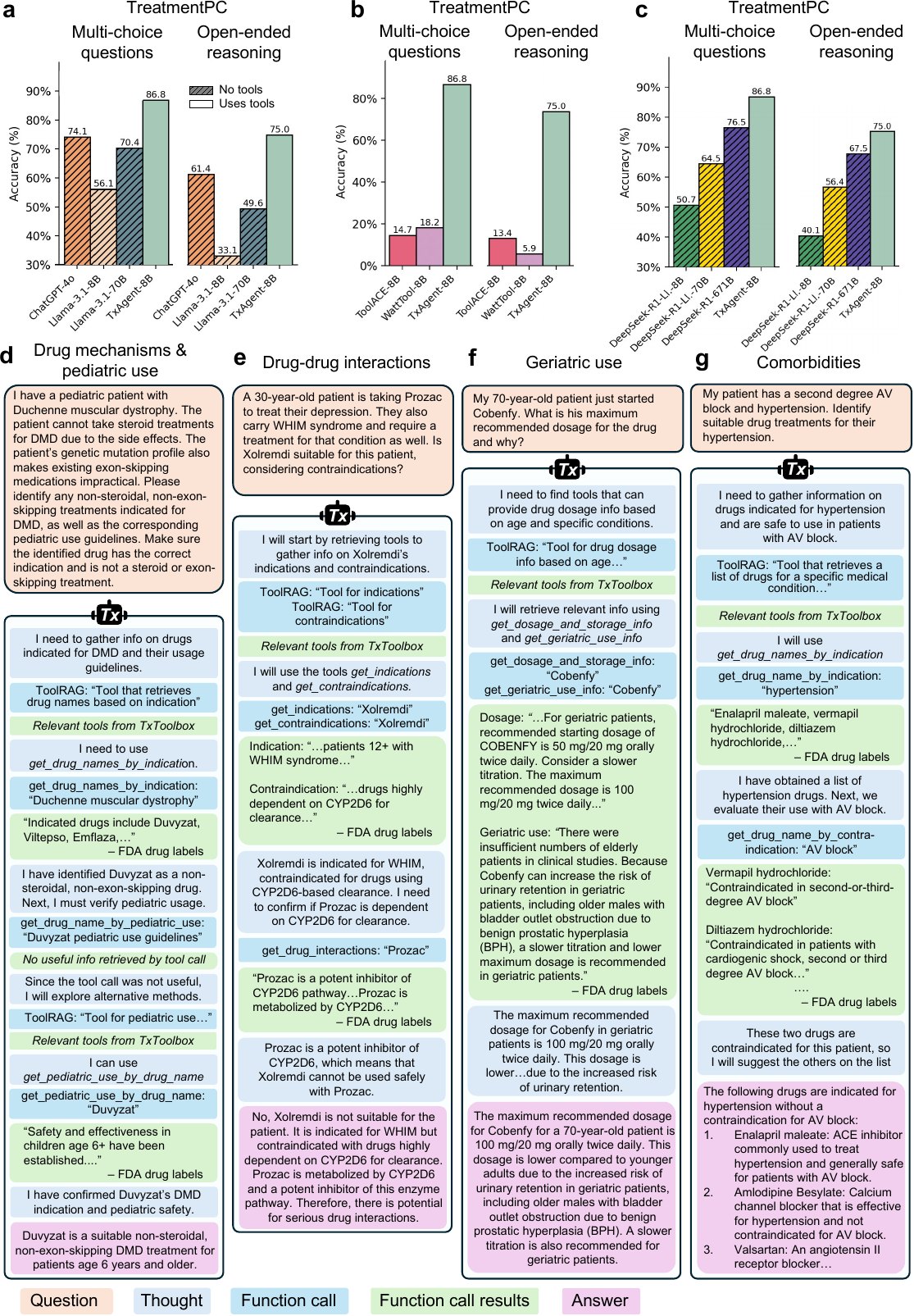}
\caption{
}
\label{fig:fig4}
\end{figure}
\clearpage
\xhdr{Figure~\ref{fig:fig4}}:
\textbf{a}) Performance comparison between \name and large-scale LLMs on the TreatmentPC benchmark. Despite being based on an 8-billion-parameter model, \name outperforms larger LLMs such as GPT-4o and Llama 3.1-70B-Instruct in both open-ended and multiple-choice settings. Notably, in the open-ended setting, \name achieves a higher accuracy (75\%) than GPT-4o does in the multiple-choice setting (74.1\%), even though the latter benefits from predefined answer options that simplify the task.
\textbf{b})~Performance comparison between \name and tool-use LLMs on the TreatmentPC benchmark.
Although ToolACE-8B and WattTool-8B, like \name, are fine-tuned on Llama-3.1-8B-Instruct and have full access to the \toolbox, \name still achieves a significantly higher performance.
\textbf{c}) Performance comparison between \name and reasoning LLMs (e.g., DeepSeek-R1) on the TreatmentPC benchmark.
\name achieves superior performance compared to the full DeepSeek-R1 model and its two distilled versions based on Llama-3.1-8B and Llama-3.3-70B.
\textbf{d}) \name identifies Duvyzat as the optimal treatment for a pediatric patient with Duchenne muscular dystrophy by evaluating drug mechanisms and pediatric use guidelines.
\textbf{e}) \name evaluates the potential drug-drug interactions between Prozac and Xolremdi, highlighting the risks of combined use due to their effects on the CYP2D6 enzyme.
\textbf{f}) \name provides personalized, evidence-based treatment advice for elderly patients, adjusting the maximum dosage of Cobenfy based on age-specific considerations and the associated risks.
\textbf{g}) \name personalizes treatment recommendations by considering comorbidities, ensuring hypertension drugs are not contraindicated for a patient's second-degree AV block.

\clearpage

\setcounter{figure}{0}
\begin{figure}[ht]
\centering
\includegraphics[width=0.9\textwidth]{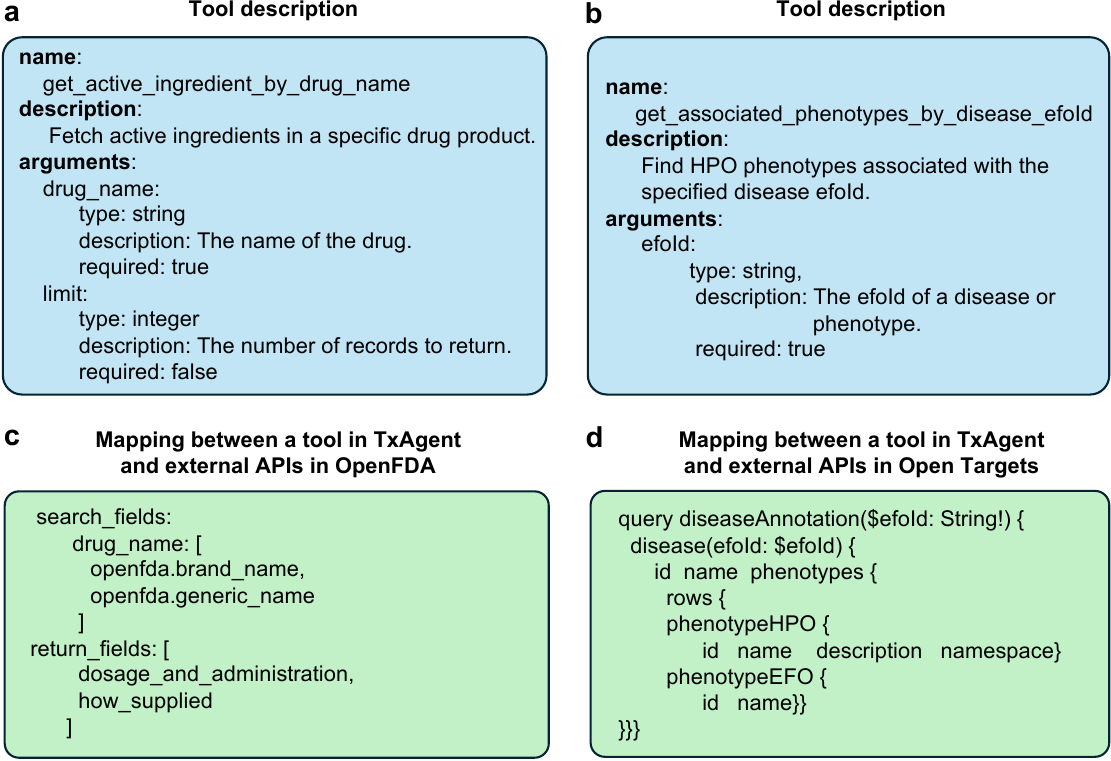}
\captionsetup{name=Extended Data Figure}
\caption{
Examples of tool specifications in \toolbox.
Each specification includes a tool description, which serves as a reference for \name's function calls, and a mapping rule that translates function calls into API requests. The tool description outlines the tool's name, purpose, and the arguments it accepts, including details such as each argument's name, purpose, data type, and whether it is mandatory.
\textbf{a}) Tool description for the tool from OpenFDA.
\textbf{b}) Tool description for the tool from Open Targets.
\textbf{c}) Mapping between Tools in TxAgent and external APIs from from OpenFDA.
\textbf{d}) Mapping between Tools in TxAgent and external APIs from from Open Targets.
}
\label{fig:tool_description}
\end{figure}

\begin{figure}[ht]
\centering
\includegraphics[width=1.0\textwidth]{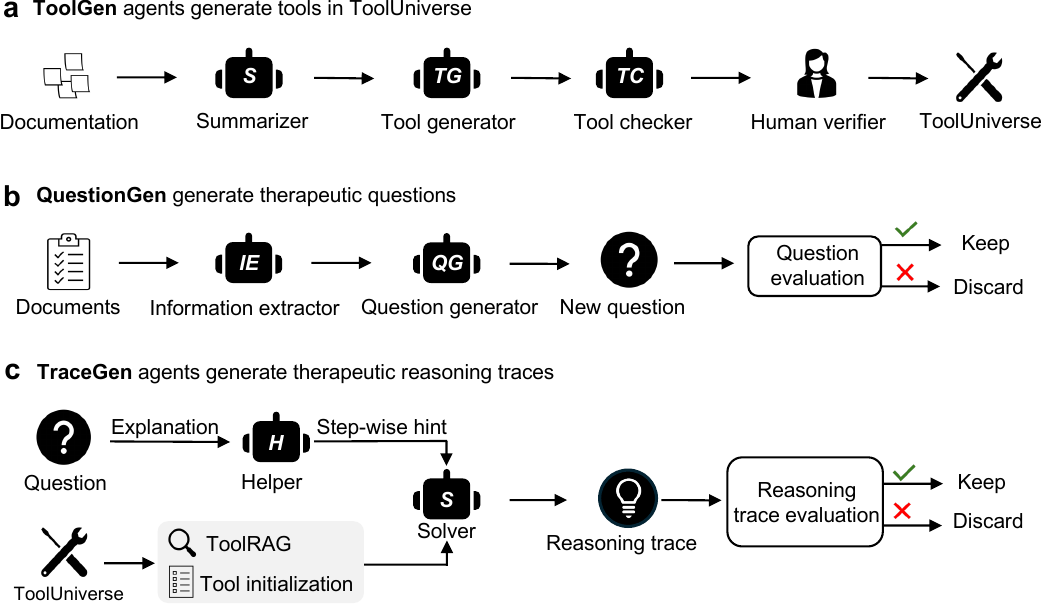}
\captionsetup{name=Extended Data Figure}
\caption{
The multi-agent systems, (i.e., \toolgen, \questgen, and \tracegen) that construct the \trainset training dataset for instruction tuning LLM to achieve the capabilities of \name.
\textbf{a}) \toolgen: A tool generation multi-agent system  that transforms APIs into 211 agent-compatible tools, aggregating them into the \toolbox.
\textbf{b}) \questgen: A question generation multi-agent system designed to extract critical information from documents (e.g., FDA drug documentation) and generate relevant questions.
\textbf{c}) \tracegen: A reasoning trace generation multi-agent system, where a \textsc{Helper} agent and a \textsc{Tool provider} module assist the \textsc{Solver} agent in generating step-by-step reasoning and function calls to solve a problem.
}
\label{fig:multi-agent-system}
\end{figure}

\begin{figure}[ht]
\centering
\includegraphics[width=1.0\textwidth]{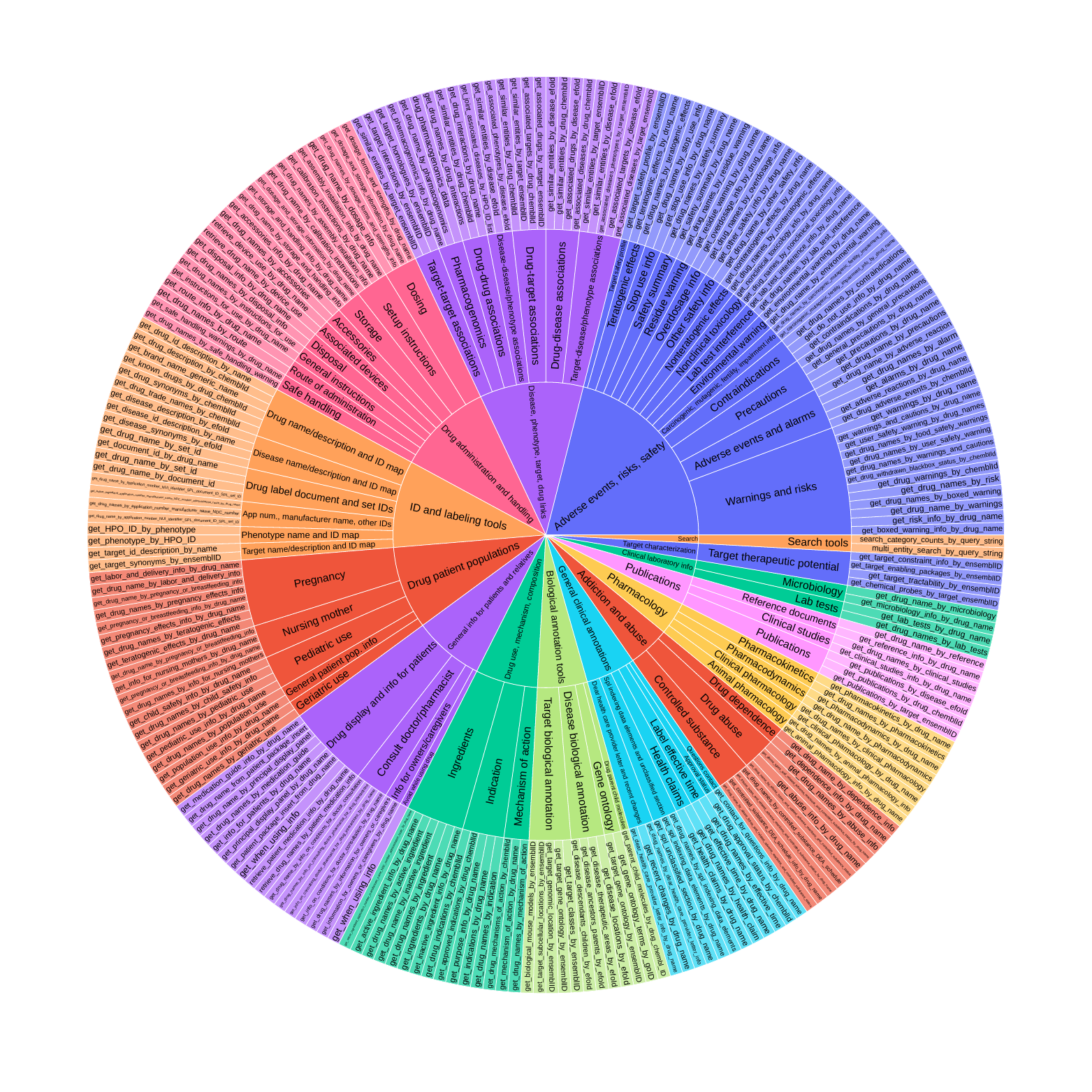}
\captionsetup{name=Extended Data Figure}
\caption{Categories of biomedical tools in \toolbox.
\toolbox contains 211 biomedical tools and includes the following categories:  
adverse events, risks, safety; addiction and abuse; drug usage in patient populations; drug administration and handling; pharmacology; drug use, mechanism, composition; id and labeling tools; general clinical annotations; clinical laboratory info; general info for patients and relatives; disease, phenotype, target, drug links; biological annotation tools; publications; search; target characterization.
}
\label{fig:tool_dist}
\end{figure}

\clearpage
\begin{figure}[ht]
\centering
\includegraphics[width=0.8\textwidth]{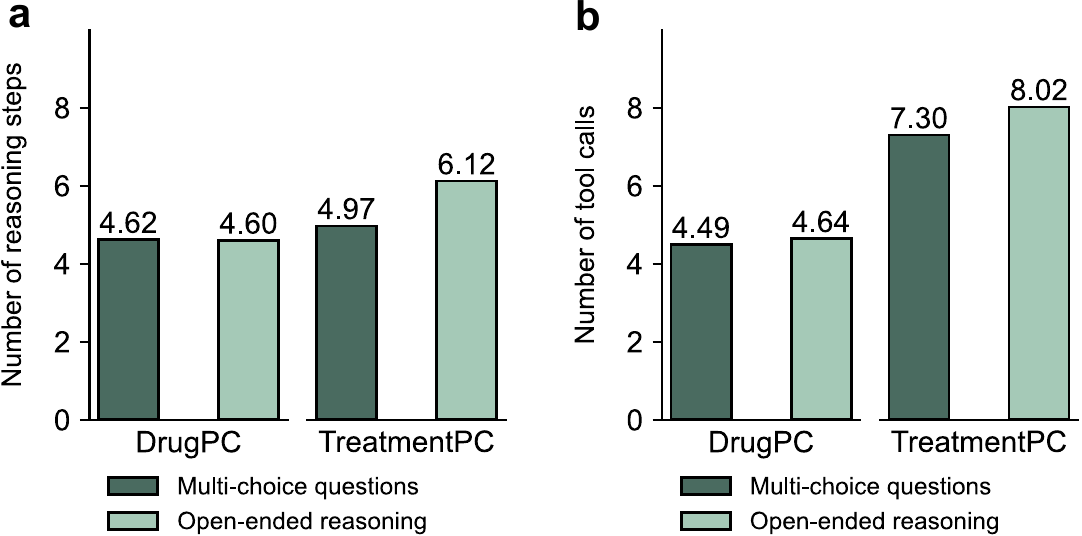}
\captionsetup{name=Extended Data Figure}
\caption{
\textbf{a}) The average number of reasoning steps for multiple-choice questions and open-ended reasoning in the DrugPC and TreatmentPC benchmarks. The TreatmentPC requires more reasoning steps compared to the DrugPC benchmarks, indicating that precision treatment recommendations require more reasoning steps before reaching a conclusion.
\textbf{b}) The average number of tool calls for multiple-choice questions and open-ended reasoning in the DrugPC and TreatmentPC benchmarks.
Similarly, the TreatmentPC benchmark requires a greater number of tool calls compared to the DrugPC.
}
\label{fig:avg_infer_step}
\end{figure}

\clearpage
\begin{figure}[ht]
\centering
\includegraphics[width=0.8\textwidth]{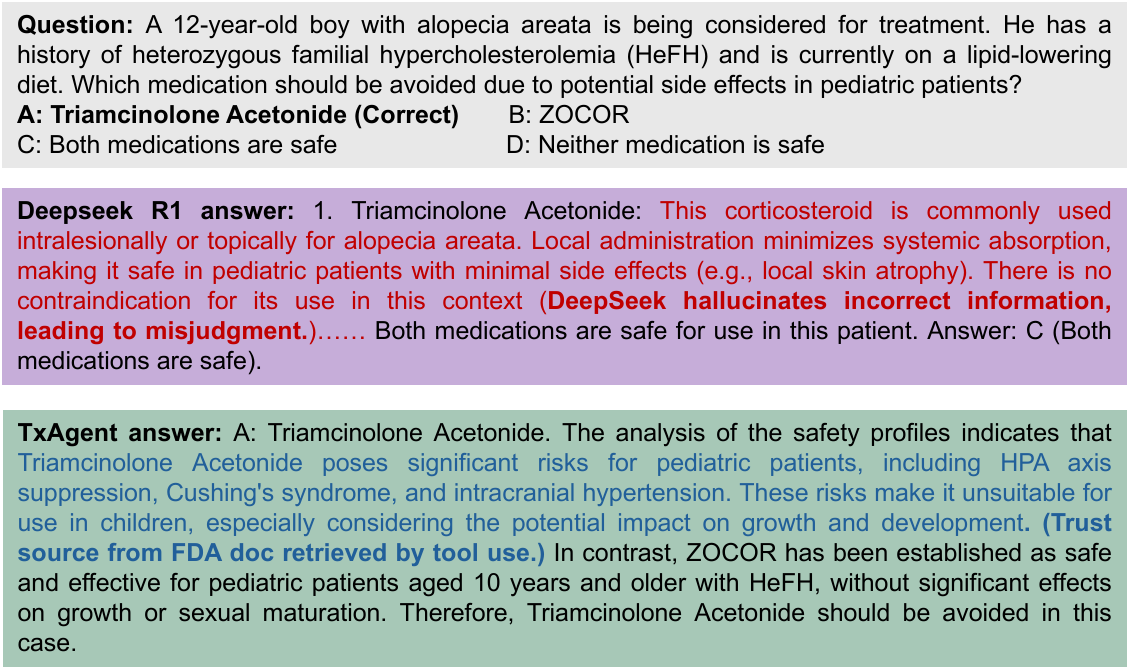}
\captionsetup{name=Extended Data Figure}
\caption{
Comparison between \name and Deepseek-R1.
Deepseek-R1 relies on its internal knowledge for reasoning, which can sometimes lead to hallucinated information and misjudgments.
In contrast, TxAgent bases its reasoning on trusted sources, such as FDA drug labels, minimizing the risk of hallucinations and ensuring more reliable conclusions.
}
\label{fig:comparison_deepseek}
\end{figure}

\clearpage
\setlength{\tabcolsep}{2mm} 
\begin{longtable}[t]{p{0.35\columnwidth}|p{0.05\columnwidth}|p{0.55\columnwidth}}
\toprule
\textbf{Benchmark}&  \textbf{Size} & \textbf{Description} \\
 \midrule
DrugPC & 3,168 & FDA newly approved drugs in 2024 \\ 
\hspace{1em} Drug Overview & 242 &  package label principal display panel; description \\
\hspace{1em} Drug Ingredients & 83 &  product data elements \\
\hspace{1em} Drug Warnings and Safety & 515 &  boxed warning; warnings and cautions; contraindications; adverse reactions; drug interactions \\
\hspace{1em} Drug Dependence and Abuse & 53 &  drug abuse and dependence; abuse; controlled substance; overdosage \\
\hspace{1em} Dosage and Administration & 507 &  indications and usage; dosage and administration; dosage forms and strengths; instructions for use \\
\hspace{1em} Drug use in Specific Populations & 333 &  use in specific populations; pregnancy; pediatric use; geriatric use; nursing mothers \\
\hspace{1em} Pharmacology & 565 &  clinical pharmacology; mechanism of action; pharmacodynamics; pharmacokinetics \\
\hspace{1em} Clinical Information & 146 &  clinical studies \\
\hspace{1em} Nonclinical Toxicology & 172 &  nonclinical toxicology; carcinogenesis and mutagenesis and impairment of fertility; animal pharmacology and or toxicology \\
\hspace{1em} Patient-Focused Information & 349 &  information for patients; patient medication guide; patient package insert; patient medication information \\
\hspace{1em} Storage and Supply Information & 203 &  how supplied; storage and handling \\
\midrule
\midrule
BrandPC & 3,168 & Drugs represented with drug brand name \\ \midrule\midrule
GenericPC & 3,168 & Drugs represented with drug generic name \\ \midrule\midrule
DescriptionPC & 626 & Drugs represented with descriptions instead of names \\  \midrule\midrule
TreatmentPC & 456 & Questions regarding specialized treatment recommendations considering patient populations \\ \midrule
\bottomrule
\caption{Benchmark datasets. These datasets are derived from newly approved FDA drugs in 2024 to minimize the risk of information leakage from LLMs. A human evaluation process is conducted to carefully review and refine the questions and answers, ensuring the exclusion of non-biomedical questions.}
\label{table:fda_11tasks}
\end{longtable}
\clearpage

\begin{table}[!t]
    \centering
\begin{footnotesize}
\setlength{\tabcolsep}{6mm} 
\begin{tabular}{cl}
\toprule
\textbf{Variable} & \textbf{Description} \\
\midrule
$Q$ & A precision therapy question. \\
$A$ & The final answer, including the rationale and the solution. \\
$G$ & The ground truth answer $G$. \\
$X$ & The explanation of why $G$ answers $Q$. \\
$\mathcal{F}$ & The \name's backend LLM. \\
$\mathcal{F}_{TX}$ & $\mathcal{F}$ prompted by the system prompt of being an \name. \\
$\mathcal{F}_{S}$ & $\mathcal{F}$ prompted by the system prompt of a summarization prompt. \\
$\mathcal{R}_i$ & The verified reasoning trace $\{R_1, R_2, ..., R_i\}$ until the $i$th step. \\
$i$ & Index representing the step in the reasoning trace. \\ 
$T_i$ & The thought at the $i$th step of the reasoning trace. \\
$ \mathcal{P}_i$ &  The collection of tools available at step $i$. \\
$\mathcal{\hat{P}}_0$ &  Tools obtained from the reference information during training data generation process. \\
$\mathcal{\hat{P}}_i^\text{RAG}$ &  Tools obtained from the \toolbox at the $i$th step during training data generation process. \\
$\mathcal{C}_i$ & The set of function calls $\{C_{i,1}, C_{i,2}, ..., C_{i,k}\}$ at step $i$. \\
$\mathcal{E}_i$ & Response of the function calls in the $i$th step. \\
$R_i$ & The $i$th step in the reasoning trace, $ R_i = [T_i, \mathcal{C}_i, \mathcal{E}_i] $. \\
$C_{i,k}$ & The $k$th function call at the $i$th step in the reasoning trace. \\
$\mathcal{B}$ & The tool set containing various tools $\{B_1, B_2, ..., B_j\}$. \\
$B_j$ & The $j$th tool from the tool set. \\
$\mathcal{I}$ & Real-world biomedical information, including FDA documents, disease and drug data. \\
\bottomrule
\end{tabular}
\end{footnotesize}
\caption{Additional notation.}
\label{tab:notation}
\end{table}
\clearpage

\begin{table*}[h!t]
\centering
	\begin{tabular}{p{0.1\columnwidth}p{0.45\columnwidth}p{0.45\columnwidth}}
		\toprule
            & Multi-choice setting & Open-ended setting\\
            \hline 
            Question & A 60-year-old female patient with a history of myocardial infarction and hypertension is being evaluated for medication adjustments. She is currently on a beta-blocker and has no known renal or hepatic impairments. Which medication would be most appropriate to add to her regimen to further reduce her cardiovascular risk? & A 60-year-old female patient with a history of myocardial infarction and hypertension is being evaluated for medication adjustments. She is currently on a beta-blocker and has no known renal or hepatic impairments. Which medication would be most appropriate to add to her regimen to further reduce her cardiovascular risk? \\
            Options &A: Sitagliptin  B: Altace C: Katerzia D: Aspirin \\
            \midrule
            Answer & B & Ramipril (Altace) is the most appropriate medication to add to the patient's regimen to further reduce her cardiovascular risk. Ramipril is an ACE inhibitor that is indicated for reducing the risk of myocardial infarction, stroke, and death from cardiovascular causes in patients 55 years or older who are at high risk of developing a major cardiovascular event. This aligns with the patient's profile, as she is 60 years old with a history of myocardial infarction and hypertension. Additionally, Ramipril is indicated for use in stable patients with signs of congestive heart failure post-myocardial infarction, making it a suitable choice for this patient. Therefore, adding Ramipril to her current regimen of a beta-blocker would provide comprehensive cardiovascular protection.\\
            \toprule
	\end{tabular}
\caption{The examples of questions, options, and answers in open-ended and multi-choice settings.}
\label{tab:example_question_type}
\end{table*}

\clearpage

\restoregeometry

\clearpage 

\spacing{1.4}
\section*{Online Methods}

\section{\name}

\subsection{Overview}
We introduce \name, an agentic AI model for precision therapy, leveraging an extensive array of biomedical tools and multi-step, white-box drug reasoning grounded with verified real-world knowledge.
\name interprets user questions written in natural language and generates answers accompanied by detailed rationales and reasoning traces (Figure~\ref{fig:fig1}b). These traces include multiple thought processes, function calls, and grounded information drawn from tools, enabling users to clearly understand and verify the basis of its conclusions.
To address complex queries, \name performs a series of actions such as analyzing user inputs and current contexts, identifying relevant tools, executing function calls on selected tools, synthesizing answers, and coordinating among tools to compile a comprehensive and accurate response.
This functionality is powered by unified multi-step reasoning, where each step involves iterative thinking and tool utilization.
\name is supported by a \toolbox and specialized tools, such as machine learning model-based tools. The \toolbox includes 211 biomedical tools for accessing high-quality knowledge, such as FDA drug information. Specialized tools are created for specific usages, such as the \toolrag, which is an embedding model that facilitates efficient tool retrieval, and the \textsc{Finish} tool, which signals the conclusion of multi-step reasoning.
The multi-step reasoning with function call abilities is achieved through fine-tuning open-source large language models (LLMs) such as Llama-3.1. By utilizing open-source models, our method supports local deployment for private applications, safeguarding patient information and ensuring privacy.

To achieve \name, the collection of \toolbox and the fine-tuning of LLMs for multi-step reasoning and function calls are essential. We introduce three key multi-agent systems: the \toolgen system for tool construction, the \questgen system for training question generation, and the \tracegen system for reasoning trace generation.
These multi-agent systems leverage AI agents powered by LLMs through prompting. Drawing on real-world biomedical information and APIs from verified sources—including FDA documents~\cite{kass2016openfda}, the OpenTarget database~\cite{ochoa2023next}, and the PrimeKG graph~\cite{chandak2022building}—they enable the generation, verification, and filtering of data. Using these systems, we construct the \trainset dataset to fine-tune LLMs and achieve \name.

\xhdr{Preliminaries.} 
During the inference process,
given a therapy question $Q$, \name generates a verified reasoning trace $\mathcal{R}_i = \{R_1, R_2, R_3, ..., R_i\}$ and the final answer $A$, which includes the rationale for the answer, such as treatment recommendations.  
The $i$-th step in the reasoning trace $R_i$ consists of a thought $T_i$, a set of function calls $\mathcal{C}_i=\{C_{i}^1, C_{i}^2, ... C_{i}^k\}$, and responses from the function calls $\mathcal{E}_{i}=\{E_{i}^1, E_{i}^2, ... E_{i}^k\}$, i.e., $R_i = \{T_i, \mathcal{C}_i, \mathcal{E}_i\}$.  
The \toolbox $\mathcal{B}=\{B_1, B_2, ..., B_j\}$ contains a wide array of biomedical tools from \toolbox.
$\mathcal{P}_i$ is the collection of tools available at step $i$,
which contains the default specialized tools and tools retrieved by the \toolrag in previous steps.
To aid understanding, Table~\ref{tab:notation} provides the complete definitions and explanations of all notations used in this work.

\subsection{Skills of \name} 
\label{sec:skill_txagent}
\name is an LLM-based agentic model designed to address complex drug reasoning problems through its versatile capabilities. These capabilities are enabled by \name's multi-step reasoning processes and its ability to perform function calls, which leverage the combined effects of its diverse skill set.
This section introduces the core skills of \name, which are obtained by the instruction finetuning of an LLM as introduced in Section~\ref{sec:training_txagent}. Then, we detail the advanced capabilities \name can achieve by integrating these skills and describe the inference process of \name.

\xhdr{Contextual thought generation.}
\name is capable of generating thoughtful, context-aware, step-by-step reasoning based on prior interactions and user inputs. Given a user query $Q$ and a sequence of previous reasoning traces $\mathcal{R}_{i-1}$, \name produces a new thought $T_i$ at step $i$, expressed  in natural language.
This thought generation process can be represented as:
\begin{equation}
T_i = \mathcal{F}_{TX}(Q, \mathcal{R}_{i-1}, \mathcal{P}_i),
\end{equation}
where $\mathcal{F}_{TX}$ is the \name's backend LLM prompted by the system prompt of being an \name, which operates in an autoregressive manner~\cite{brown2020language}, and $\mathcal{P}_i$ is a set of available tools at this step.
The query $Q$ along with the reasoning traces $\mathcal{R}_{i-1}$ is provided as input to $\mathcal{F}_{TX}$ to generate $T_i$, which incorporates reasoning about the analysis of prior steps and determines the next actions.

\xhdr{Function call arguments generation.}
\name executes tools by generating function call arguments based on the tool descriptions provided in the prompts. Each tool description consists of the tool's name, its purpose, and the arguments it accepts. For each argument, the description specifies its name, purpose, data type, and whether it is mandatory (\extfigref{fig:tool_description}).
Following the generated thought $T_i$ of reasoning step $R_i$, given a set of tool descriptions $\mathcal{P}_i$ available to \name, \name produces the corresponding function call arguments for tool calls: 
\begin{equation}
\mathcal{C}_i  = \mathcal{F}_{TX}(Q, \mathcal{R}_{i-1}, T_i, \mathcal{P}_i),
\end{equation}
where $\mathcal{C}_i = \{C_{i}^1, C_{i}^2, \dots, C_{i}^{k}\}$ is a list that contains multiple function calls as \name supports parallel tool execution by generating multiple function calls across various tools.
The $k$-th function call at step $i$, $C_{i}^{k}$, is represented as a code snip in JSON format:
\begin{equation}
C_i^k = \left\langle \texttt{name}, \mathcal{A}_i^k \right\rangle,
\end{equation}
where $\texttt{name}$ is the name of the tool selected from the set of available tools $\mathcal{P}_i$, $\mathcal{A}_i^k = \{a_1, a_2, \dots, a_n\}$ represents the arguments required by tool $C_i^k$, where each $a_j$ corresponds to a specific argument-value pair corresponding to the description of the selected tool.
The generated function call arguments $\mathcal{C}_i$ are sent to the \toolbox codebase for execution. The results from these function calls $\mathcal{E}_{i}=\{E_{i}^1, E_{i}^2, ... E_{i}^k\}$, where $E_{i}^k$ is the result of $k$-th function call, are then sent back to \name as part of the current step of the reasoning trace $R_i = \{T_i, \mathcal{C}_i, \mathcal{E}_i\}$.
The reasoning trace $ \mathcal{R}_i $ is then updated as: $\mathcal{R}_i = \mathcal{R}_{i-1} \cup R_i$.
Tools used by \name involve a variety of types, such as biomedical tools in \toolbox that gather outputs from multiple verified sources, and the machine learning-based tools (\toolrag) that use a machine learning model to achieve certain functions.
Moreover, we will explore tool use of \name in a broader range,  
such as the formation of a multi-agent system where tools represent specialized agents with unique capabilities, or even the creation of a new instance of \name. To add new capabilities to \name, simply introduce new tools to the framework, and \name will automatically incorporate and utilize them when needed.

\xhdr{Logical multi-step reasoning and decision-making.}
\name achieves logical multi-step reasoning by iteratively generating thoughts and formulating tool arguments. At each step, \name evaluates whether the reasoning trace up to that point—particularly the outputs from prior function calls—provides sufficient information to answer the user's query by generating thought $T_i$. Based on this assessment, it decides on the next action, which could involve generating new function calls $\mathcal{C}_i$ to use tools for accessing new information or generating the final answer $A$, depending on if  the special token \texttt{[FinalAnswer]} is generated or not:
\begin{equation}
\begin{aligned}
\mathcal{C}_i = \mathcal{F}_{TX}(Q, \mathcal{R}_{i-1}, T_i, \mathcal{P}_i) , \quad & \texttt{[FinalAnswer]} \notin T_i; \\
A, \mathcal{C}_{F}  = \mathcal{F}_{TX}(Q, \mathcal{R}_{i-1}, T_i, \mathcal{P}_i) , \quad & \texttt{[FinalAnswer]} \in T_i,
\end{aligned}
\end{equation}
where $\mathcal{C}_{F}$ is a final function call to the special \textsc{Finish} tool, signifying the end of the reasoning process.
Despite its simplicity, this iterative thought and function call generation process has potential to incorporate complex and dynamic agentic workflows.

\xhdr{Proactive tool search, selection and utilization.}
\toolbox includes 211 tools spanning various aspects of the biomedical field. However, due to the limited context window of the LLM (i.e., the number of text tokens it can handle in a prompt), it is impractical to include all tool descriptions within the prompt.
To address this, \name employs a proactive tool search strategy by utilizing the \toolrag. When no suitable tool is available for the next action, \name dynamically invokes the \toolrag by function call to search for tools matching the desired requirement.
Rather than relying on tools memorized during training, \name selects and uses tools based on its current requirements and descriptions of tools, ensuring flexibility and scalability.
The \toolrag is an embedding model designed to retrieve tools based on specific requirements.
During inference, the \toolrag processes all tool descriptions in \toolbox to generate their semantic embeddings. For each new function call to \toolrag, the model encodes the requirement argument obtained from the function call arguments into an embedding and retrieves the top-$k$ tools whose embeddings have the highest similarity to the requirement's embedding.
The newly retrieved tools are put into the tool set $\mathcal{P}_i$.
Due to the expansive scope of \toolbox, imperfection of \toolrag, and the open-ended reasoning approach of \name, not all tools retrieved by \toolrag are immediately applicable to \name's next action. Drawing from the prior reasoning trace and the descriptions of the retrieved tools, \name identifies the most suitable options from $\mathcal{P}_i$, formulates new thoughts, and generates parallel function calls to effectively utilize the selected tools.

\xhdr{Concise summarization.}
Tool outputs can often be lengthy, especially in complex cases. This poses a challenge due to the limited context window of the LLM, as lengthy outputs restrict the maximum number of reasoning steps that can be performed. To overcome this limitation, \name introduces a mechanism to transform lengthy tool outputs into concise, accurate, and meaningful summaries.
Given a reasoning step $R_i = \{T_i, \mathcal{C}_i, \mathcal{E}_i\}$, \name generates a summarized version of tool response $\mathcal{E}_i$ as follows:
\begin{equation}
\hat{\mathcal{E}}_i = \mathcal{F}_{S} \left( T_i, \mathcal{C}_i, \mathcal{E}_i \right), 
\end{equation}
where $\mathcal{F}_{S}$ is the backend LLM of \name prompted with a summarization prompt.
This summary retains the essential information relevant to the thought $T_i$, ensuring that the critical details are preserved. By compressing lengthy outputs into a compact form, \name enables a larger number of reasoning steps while avoiding context window overflow.

\xhdr{Structured question responses.}
While \name generates open-ended answers along with a reasoning trace, it can also be used for evaluating multiple-choice questions. Given a question and the open-ended answer, \name can map the answer to the correct option from the provided choices.

\begin{algorithm}[th]
\caption{\name multi-step inference process.}
\label{alg:txagent_inference}
\KwIn{Question $Q$, \toolbox $\mathcal{B}$, Initial available tools $\mathcal{P}_0$}
\KwOut{Reasoning trace $\mathcal{R}$, final answer $A$}

Initialize $\mathcal{R} \gets \{\}$, tools $\mathcal{P} \gets \mathcal{P}_0$, step $i \gets 0$ \;

\While{Reasoning is incomplete}{
    $i \gets i + 1$\;

    Generate thought: $T_i = \mathcal{F}_{TX}(Q, \mathcal{R}_{i-1}, \mathcal{P}_i)$ ; \\
    
    \If{\texttt{[FinalAnswer]} in $T_i$}{
        Generate final answer: $A, \mathcal{C}_{F}  = \mathcal{F}_{TX}(Q, \mathcal{R}_{i-1}, T_i, \mathcal{P}_i)$; \\
        Execute \textsc{Finish} tool to end the multi-step reasoning; \\
        \textbf{Return} $\mathcal{R}_i, A$\;
    }
    \Else{
    Generate function calls: $\mathcal{C}_i  = \mathcal{F}_{TX}(Q, \mathcal{R}_{i-1}, T_i, \mathcal{P}_i)$; \\
    \If{call to \textsc{ToolRAG} in $\mathcal{C}_i$}{
        Execute \textsc{ToolRAG} and update $\mathcal{P}_i$; } 
    \Else{
    Execute tools from $\mathcal{C}_i$ to get tool response $\mathcal{E}_i$;
    }
    Update reasoning trace: $\mathcal{R}_i \gets \mathcal{R}_{i-1} \cup \{T_i, \mathcal{C}_i, \mathcal{E}_i\}$\;
    }
}
\end{algorithm}

\subsection{Capabilities of \name} 
\label{sec:cap_txagent}
In Algorithm \ref{alg:txagent_inference}, we present the inference process of \name, leveraging the skills described above. We highlight the capabilities of \name made possible through its diverse skill sets.

\xhdr{Knowledge grounding using tool calls.}
The treatment problem demands reliable answers accompanied by transparent explanations to justify decisions. However, a significant concern arises from the inability of machine learning models, such as LLMs, to provide dependable explanations for their predictions. This forces users to invest additional effort in determining whether the model's predictions can be trusted.
With the function calling skill, \name provides answers to user queries grounded in verified information by leveraging tools connected to trusted sources. Instead of generating responses directly like traditional LLMs, \name utilizes tools to retrieve accurate information. The answers are then crafted based on the verified outputs of these tools. For instance, it can query the dosage instructions for a medication from official FDA documents. 
This knowledge-grounding approach allows users to validate the factual correctness of answers by reviewing the reasoning trace, ensuring transparency and reliability.

\xhdr{Goal-oriented tool selection.}
Through the proactive tool search, selection and utilization skills, \name leverages \toolrag to search for tools, identify suitable options, and effectively utilize the most appropriate tools from the candidates provided by \toolrag.
This approach enables \name to access a vast array of tools and seamlessly adapt to new ones.
Instead of relying solely on tools memorized during training, the goal-oriented tool selection process allows \name to reason more freely by first generating a plan of action and then identifying the tools necessary to execute it.
Furthermore, \name can expand its capabilities by integrating additional tools into the \toolbox without requiring retraining. When faced with new scenarios where existing tools are insufficient, \name can address these cases by incorporating relevant tools into the \toolbox, showcasing its flexibility and adaptability in handling novel challenges.

\xhdr{Multi-step therapeutic reasoning.}
When tackling complex therapeutic problems that cannot be solved in a single step, \name employs multi-step therapeutic reasoning to iteratively generate new thoughts and function calls based on the prior reasoning trace.
There are two key scenarios where multi-step reasoning is essential. First, solving complex problems often requires gathering information from multiple perspectives before arriving at a well-founded answer. Second, in real-world applications, interactions with the environment can be unpredictable—such as function calls failing to retrieve necessary information—making it common for a single attempt to fall short.
By leveraging multi-step reasoning, \name can effectively address both cases by systematically collecting information, generating new ideas, and making additional function calls to explore alternative solutions. This iterative process continues until the goal is successfully achieved.

\xhdr{Real-time retrieval from continually updated knowledge sources.}
Once a model finishes training, its internal knowledge remains static and is no longer updated. Given the high cost and technical challenges associated with continuously training large models, such as LLMs, it is difficult to incorporate new knowledge directly into these models.
Retrieval-augmented generation \cite{lewis2020retrieval}, a special form of tool-use model, retrieves relevant text by matching query embeddings with a precomputed vector database. However, maintaining a high-quality vector database is computationally intensive, making frequent updates difficult.
\name takes a different approach by using function calls to directly access multiple constantly updated data sources, such as the OpenTargets and FDA databases. 
By leveraging these dynamic knowledge bases, \name can answer questions about newly approved drugs, even when the training data lacks relevant information.
Additionally, it integrates complementary information from multiple sources, eliminating the need to construct and maintain a vector database.

\section{\toolbox}
The \toolbox consists of 211 biomedical tools that provide real-time, up-to-date information on diseases, drugs, targets, and other essential biomedical data. 
Constructing such a vast number of tools manually would be impractical; therefore, we developed \toolgen, a tool construction multi-agent system that automates the creation of tool descriptions and the critical mappings between tools and APIs.
This section first presents an overview of \name, followed by an introduction to \toolgen system.

\subsection{Overview of \toolbox}
\toolbox has 211 biomedical tools, covering the following categories: 
adverse events, risks, safety; addiction and abuse; drug patient populations; drug administration and handling; pharmacology; drug use, mechanism, composition; ID and labeling tools; general clinical annotations; clinical laboratory info; general info for patients and relatives; disease, phenotype, target, drug links; biological annotation tools; publications; search; target characterization.
Tools in \toolbox are built upon APIs from multiple sources, including OpenFDA, OpenTargets, and the Monarch Initiative.
The complete distribution of tools across these categories is presented in \extfigref{fig:tool_dist}.
Each tool includes a tool description that will be provided to \name as the reference for function call, along with backend code that translates function calls into API requests to these external sources. 
Tool description consists of the tool's name, its purpose, and the arguments it accepts. For each argument, the description specifies its name, purpose, data type, and whether it is mandatory (Examples in \extfigref{fig:tool_description}).

\subsection{\toolgen: a multi-agent system for constructing tools}
\label{sec:data_gen_method}
\toolgen is a tool construction multi-agent system for constructing tools that are suitable for \name, based on API documentation.
API documentation often varies significantly in format and content, presenting challenges for direct integration into \name. For instance, OpenTargets utilizes a GraphQL schema to describe its API; OpenFDA employs Elasticsearch as its API backend and includes documentation to explain available fields; The Monarch Initiative provides RESTful APIs.
This diversity in API representation complicates the process of converting them into tools for \name.
\toolgen system addresses this by organizing the API functions into a set of tools, each with a specific purpose and a clear description that is easily understandable to \name.
\toolgen system comprises three agents: the \textsc{summarizer}, \textsc{tool generator}, and \textsc{tool checker} (\extfigref{fig:multi-agent-system}a). Since their abilities are simple, these agents are implemented by providing specialized instructive prompts to GPT-4o.
After completing the tool construction, a human evaluation process is conducted to assess the generated tools.

\xhdr{API summarization.} The \textsc{summarizer} agent serves as the initial step in the system. It extracts API documentation from a given source to summarize the API's capabilities. The result is a list of potential functions that could be enabled using the APIs, such as ``identify the active ingredients for a drug" and ``find disease-related phenotypes".

\xhdr{Tool construction.} For each capability in the list, the \textsc{tool generator} agent refers to the API documentation to create detailed tool specifications. These specifications include the tool's name, description, arguments, and specialized mapping data to translate arguments into API requests.
Each argument includes the name, description, data type, and an indication of whether it is required.
Mappings for OpenTargets and Monarch Initiative correspond to the query string defined in the GraphQL and RESTful schema, with variables in APIs connected to arguments in the tool. Mapping for OpenFDA employs the search and return fields of Elasticsearch, where search fields are tied to arguments in the tool, and return fields are selected to align with the tool's description (\extfigref{fig:tool_description}).

\xhdr{Tool check.}  The \textsc{tool checker} agent evaluates the validity of generated tools by constructing and testing questions and function calls. In this process, we first verify the mapping, ensuring its correctness by checking the validity of the provided mappings. Next, we randomly sample data points linked to either the input or output of the APIs, such as drug names, disease names, target names, and their corresponding IDs, to test the APIs.
If useful information can be retrieved through API requests, the retrieved data and the tool specifications are sent to the \textsc{tool checker} agent. The agent then generates test questions and function call arguments for the tool. If the generated function call arguments produce valid outputs, the tool is deemed functional and valid.
However, if any of the steps in this process fail, the tool is marked invalid and subsequently removed.

\xhdr{Human verification.}  After the tool construction process, human experts manually verify and refine the tools. This evaluation includes determining whether the tool has meaningful applications, verifying that it functions as described, and ensuring its stability when handling unexpected inputs. Once this process is complete, the validated tools are included in the \toolbox. 

\subsection{Tool Graph}
The tool graph is a directed graph that connects tools in \toolbox. It is used to facilitate the construction of training data. In the tool graph, each node represents a tool, and a directed link is established when the output of one tool serves as the input for another. The presence of a link is determined by providing descriptions of the two tools to an LLM, which then decides if a directed link should exist between them.
Sampling a tool chain from a tool graph enables the construction of complex questions that require multiple rounds of tool calls. Further uses of the tool graph are described in Section~\ref{sec:train_set}.
The tool graph is used solely for constructing the training dataset, not for the inference process in \name, due to the challenges in constructing an exceptionally precise tool graph. Unrestricted by the tool graph, \name can seamlessly integrate newly added tools during the inference process.

\section{Constructing \trainset dataset}
\label{sec:train_set}
We perform instruction tuning on open-source LLMs using a collected \trainset dataset to achieve the capabilities of \name. This section describes the construction process of \trainset.
To achieve comprehensive coverage of specialized treatment and drug information, we employ \questgen, a question construction multi-agent system that produces diverse questions.
Recognizing the challenges in generating valid reasoning traces that effectively integrate feedback from real-world tools, we design \tracegen, a multi-agent system that leverages a helper agent to assist in generating complex, step-wise reasoning traces.

\subsection{\trainset Data Sources}
\label{label:data_source}
The source information of \trainset is collected from the following sources.

\xhdr{OpenFDA} is a health informatics database maintained by the FDA, offering public access to FDA data on approved drugs, devices, and foods~\cite{kass2016openfda}. This work utilizes the drug labeling data provided by the platform, which covers more than 67,000 drugs currently on the market~\cite{openfda}. OpenFDA includes a search API for retrieving information based on specific query fields.
In this study, we use the drug API of OpenFDA to obtain FDA documentation on various drugs. Each drug entry contains numerous fields, such as indications, boxed warnings, and supply information (Table~\ref{table:fda_11tasks}).

\xhdr{Open Targets} is a platform that integrates data from 23 public resources, including Orphanet, Gene2Phenotype, and ChEMBL~\cite{ochoa2023next}, to facilitate target identification and prioritization. As of September 2024, Open Targets includes 63,121 targets, 28,327 diseases, 18,041 drugs, 17,853,184 evidence entries, and 8,155,988 target-disease associations~\cite{opentargets2024}.
In this study, we utilize the association data from Open Targets to extract drug-disease relationships, drug status, drug-target interactions, and disease-target associations.

\xhdr{Human Phenotype Ontology from the Monarch Initiative} (HPO) is a database that provides an ontology of medically relevant phenotypes and disease-phenotype annotations~\cite{castellanos2024human}. It includes over 18,000 terms and more than 156,000 annotations linked to hereditary diseases. We leverage HPO to establish connections between diseases and phenotypes.

\xhdr{PrimeKG} is a comprehensive medicine-focused knowledge graph designed to offer a holistic view on diseases~\cite{chandak2022building}. It incorporates data from 20 high-quality biomedical resources, capturing details about 17,080 diseases and their 4,050,249 relationships across ten key biological scales. In this work, PrimeKG provides the disease list and disease-related information for generating disease-related questions.

\subsection{\questgen Multi-agent System for Question Construction}
\label{sec:question_gen}
While training \name requires a large number of diverse questions that cover different aspects regarding treatment, disease, and drugs, and considers specialized cases such as patient populations, drug side effects, and drug interactions, manually writing these questions would be too costly.
To effectively collect questions, \questgen system is proposed as a question construction multi-agent system that generates meaningful questions from verified knowledge bases such as FDA documents.
\questgen system begins with the information extraction, which identifies and extracts key information relevant to the desired questions from documents and data sources. Using the extracted information, the question construction step creates questions, corresponding answers, and detailed explanations that clarify how the answer addresses the question.
At last, the question evaluation step verifies questions in multiple aspects.

\xhdr{Question types.} We generate questions through three distinct approaches: drug-centered, disease-centered, and tool-chain-centered question construction.
Drug-centered questions focus on common therapeutic aspects of drugs, including their use in specific patient populations, indications, dosage, safety warnings, and potential risks.
Disease-centered questions address specialized treatment scenarios. These questions incorporate detailed patient profiles, such as phenotypes, medical histories, current medications, and characteristics of the patient population.
Tool-chain-centered questions are generated by randomly sampling a sequence of tools from the \toolbox tool graph, followed by creating questions based on the selected tool chain, which increases the diversity of questions.

\xhdr{Information extraction.}
This step identifies and extracts key information relevant to the desired questions from documents and data sources.
Different information extraction strategies are designed to meet the requirements of various question types. 

For drug-centered questions, we randomly sample drugs from the FDA database and retrieve their corresponding FDA documents as raw data sources. From each drug's FDA document, one field is randomly selected and extracted as the reference data for question construction. To enable question construction beyond just related to drug names, descriptive information about the drug—such as its mechanism of action, indications, and contraindications—is also extracted. This allows for the creation of questions that focus on the drug's characteristics without explicitly mentioning its name.

For disease-centered questions, we begin by randomly sampling a disease and gathering its description, associated phenotypes, targets, and all potential drugs. For each drug in the list, we retrieve its FDA document and extract information on indications, patient populations, contraindications, warnings, and drug interactions.
The extracted data is then categorized by field and passed to the \textsc{information extractor} agent, which compares the drugs across these fields and highlights their differences. The generated comparison serves as the reference for question construction, enabling the creation of challenging, specialized questions that account for subtle differences among drugs.

For tool-chain-centered questions, we first sample a tool-chain from the tool graph starting from common tools such as identify drug ID or disease ID based on names.
Then, we obtain information that can be retrieved by tools and the tool descriptions as the reference for question construction.

\xhdr{Question construction.}
In this step, the \textsc{Question Generator} leverages reference information extracted during the information extraction process to produce the question $Q$, corresponding answer $G$, and explanations justifying why the answers are correct $X$. For multiple-choice questions, it also generates answer options.
The inclusion of explanations plays a crucial role, as they ensure the meaningfulness of the generated questions and offer solution hints for the \textsc{Helper} agent during reasoning trace generation.

The \textsc{Question Generator} operates by prompting GPT-4o with instructions for generating questions. It utilizes multiple prompt variations, each designed for specific question types. During the question-construction process, general guidelines for question creation, reference information, and specific requirements for particular question types are provided as contextual input to the \textsc{Question Generator} to produce the questions.

\xhdr{Question evaluation.}
The generated question undergoes evaluation based on three key aspects: knowledge-based grounding, answerability, and reasonableness. For each aspect, GPT-4o is prompted to perform the evaluation.
For knowledge-based grounding, to ensure the question is generated from the reference information and not from hallucinations by the language model, both the question and reference information are provided to GPT-4o. GPT-4o is tasked with verifying whether the information in the question is directly derived from the reference information.
For the answerability check, GPT-4o is prompted to assess whether the question can be adequately answered using the provided reference information.
For the reasonableness check, the explanation in the generated question is sent to GPT-4o, which evaluates whether the reasoning behind the explanation is logical and makes sense.
If any of the checks fail, the question is discarded. Otherwise, it is retained and sent to \tracegen for reasoning trace construction.

\subsection{Reasoning Trace Generation}
\label{sec:trace_gen}
\tracegen is designed to generate training data consisting of a reasoning trace $\mathcal{R}$ and the final answer $A$ based on the question $Q$. However, generating $\mathcal{R}$ faces several challenges:
1) Complexity of questions: Many questions require multi-step reasoning and analysis of multiple aspects, making it difficult to generate a single straightforward answer. The challenge is to create a reasoning trace that can handle these complexities effectively.
2) Incorporating external tools: To improve reasoning with the help of a massive number of tools, it's important to incorporate real-world tools into $\mathcal{R}$. The challenge here is integrating the outputs of these tools, rather than relying solely on the internal knowledge of LLMs.
3) Handling uncontrollable tool outputs: The results from external tools are often unpredictable. A key challenge is how to manage failure cases and continue progressing toward a solution, even when tool outputs deviate from expectations.

\tracegen is a multi-agent system designed to address various challenges through its key components: the \textsc{Helper} agent, the \textsc{Tool Provider} module, and the \textsc{Solver} agent.
The \textsc{Helper} agent assists the \textsc{Solver} by offering step-by-step solution hints. It has access to the answers and explanations for questions and provides guidance for the next steps in the reasoning process based on the prior steps generated by the \textsc{Solver} agent.
The \textsc{Tool Provider} presents a selection of potential tools for the \textsc{Solver} agent to choose from. These tools are identified based on reference information from the current question and a \toolrag, which is iteratively trained on previously collected data to improve its recommendations.
Armed with tools from the \textsc{Tool Provider} module, hints from the \textsc{Helper} agent, the current question, and previously generated reasoning traces, the \textsc{Solver} agent iteratively solves the problem. It does so by generating subsequent reasoning steps and function calls until arriving at the final answer.

\xhdr{Providing solution hint with \textsc{Helper}.}  
The \textsc{Helper} agent plays a crucial role in assisting the \textsc{Solver} by providing solution hints, which is achieved by prompting GPT-4o with instructions. At each reasoning step $i$, the \textsc{Helper} has access to the problem question $Q$, the ground truth answer $G$, and its explanation $X$. Additionally, it takes as input the current reasoning trace $\mathcal{R}_i = \{R_1, R_2, ..., R_i\}$, which represents all steps generated by the \textsc{Solver} up to step $i$. Using this information, the \textsc{Helper} generates a solution hint, denoted as $\mathcal{H}_{i+1}$, which guides the \textsc{Solver} toward the next step in the reasoning process.
When the \textsc{Solver} provides an answer $A$, the \textsc{Helper} checks whether $A$ matches the ground truth answer $G$. If $A = G$, the reasoning process is deemed complete. If $A \neq G$, the \textsc{Helper} prompts the \textsc{Solver} to reflect on its reasoning and continue the reasoning process. In such cases, the \textsc{Helper} generates a hint $\mathcal{H}_{i+1}$ to guide the \textsc{Solver} back into the reasoning process and help refine the answer.
\textsc{Helper} is defined as:
\begin{equation}
\mathcal{H}_{i+1} = \textsc{Helper}([Q, G, X], \mathcal{R}_i, A),
\end{equation}
where $A$ is empty if it is not provided to the \textsc{Helper}.
By iteratively providing hints $\mathcal{H}_{i+1}$, the \textsc{Helper} ensures that the \textsc{Solver} progresses logically, generating the reasoning trace until the solution $A$ is fully constructed and consistent with the ground truth answer $G$ and explanation $X$.

\xhdr{Providing tools with the \textsc{Tool Provider}.}  
The \textsc{Tool Provider} module supports the \textsc{Solver} by supplying relevant tools during the reasoning process. It operates in two stages. First, the module analyzes the reference information attached to the problem question $Q$ and identifies an initial set of tools $\mathcal{\hat{P}}_0$ from the tool set $\mathcal{B}$. These initial tools are provided to the \textsc{Solver} at the start of the reasoning process. Second, if the \textsc{Solver} determines that no suitable tools are available for a specific reasoning step, it invokes the \textsc{ToolRAG} model within the \textsc{Tool Provider} module. This model retrieves additional tool suggestions, denoted as $\mathcal{\hat{P}}_i^\text{RAG}$, based on the tool descriptions provided by the \textsc{Solver}. 
By combining these two stages, the \textsc{Tool Provider} module ensures that the \textsc{Solver} has access to the most relevant tools throughout the reasoning process, either by leveraging the initial set of tools $\mathcal{\hat{P}}_0$ or dynamically adapting to the problem's demands with tools $\mathcal{\hat{P}}_i^\text{RAG}$ from the \textsc{ToolRAG} model.

\xhdr{Step-wise reasoning trace generation with \textsc{Solver}.}  
The \textsc{Solver} serves as the central component for iteratively generating the reasoning trace $\mathcal{R}$ and deriving the final answer $A$, which is achieved by prompting GPT-4o. We provide the algorithm in~Algorithm~\ref{alg:solver_trace}. At each step $i$, the \textsc{Solver} integrates the question $Q$, tools from the \textsc{Tool Provider}, solution hints $\mathcal{H}_{i}$ provided by the \textsc{Helper}, and its prior reasoning $\mathcal{R}_{i-1}$. Using this information, it formulates intermediate thoughts $T_i$ and function calls $\mathcal{C}_i$, driving the reasoning process toward a complete and accurate solution.
To simulate the inference process, the \textsc{Solver} avoids directly utilizing tools available in the initial set $\mathcal{\hat{P}}_0$. Instead, it generates virtual \textsc{ToolRAG} calls, which simulate accessing these tools. Each virtual call specifies the tool’s name and its rewritten description from $\mathcal{\hat{P}}_0$. These virtual calls are later replaced by actual calls to \textsc{ToolRAG}.
When the \textsc{Solver} identifies that no suitable tools are available for the current reasoning step, it invokes the \textsc{ToolRAG} model within the \textsc{Tool Provider} to dynamically suggest additional tools. These new tools, denoted as $\mathcal{\hat{P}}_i^\text{RAG}$, are retrieved based on descriptions provided by the \textsc{Solver}.
Hints $\mathcal{H}_{i+1}$ from the \textsc{Helper} guide the \textsc{Solver} through the reasoning trajectory by suggesting logical next steps. This iterative mechanism allows the reasoning trace to evolve through external tool usage, dynamic adjustments based on feedback, and updates to the reasoning trace. The process continues until the \textsc{End} tool suggests a candidate answer $A$. If validated by the \textsc{Helper}, the reasoning trace $\mathcal{R}$ and the final answer $A$ are returned. If the answer is deemed incorrect, the \textsc{Solver} removes the corresponding reasoning step and continues refining $\mathcal{R}$.

\begin{algorithm}[H]
\caption{Step-wise reasoning trace generation with \textsc{Solver}}
\label{alg:solver_trace}
\KwIn{Question $Q$, \toolbox $\mathcal{B}$, ground truth $G$, explanation $X$}
\KwOut{Reasoning trace $\mathcal{R}$, final answer $A$}

Initialize $\mathcal{R} \gets \{\}$, tools $\mathcal{P} \gets \{\}$,  step $i \gets 0$ \;
Obtain initial hints $\mathcal{H}_{0}$ from \textsc{Helper}\;
Obtain initial tools $\mathcal{\hat{P}}_0$ from \textsc{Tool Provider}\;

\While{Reasoning is incomplete}{
    $i \gets i + 1$\;
    \If{Suitable tools exist in $\mathcal{P}$}{
        Generate thought $T_i$ and call $\mathcal{C}_i$ based on $Q$, $\mathcal{H}_{i}$, and $\mathcal{R}_{i-1}$\;
    }
    \If{Suitable tools exist in $\mathcal{\hat{P}}_0$}{
        Generate thought $T_i$ and virtual calls $\mathcal{C}_i$;
        \ForEach{Virtual call in $\mathcal{C}_i$}{
            Replace virtual \textsc{ToolRAG} call with real arguments;
        }
    }

    \If{No suitable tools in $\mathcal{P}$}{
        Generate thought $T_i$ and request additional tools $\mathcal{\hat{P}}_i^\text{RAG}$ by calling \textsc{ToolRAG} with desired tool's descriptions\;
    }
    
    Execute tool calls from $\mathcal{C}_i$ and update reasoning trace: $\mathcal{R} \gets \mathcal{R} \cup \{R_i\}$\;
    Obtain the next hint $\mathcal{H}_{i+1}$ from \textsc{Helper}\;
    
    \If{\textsc{End} tool provides candidate answer $A$}{
        \If{\textsc{Helper} confirms correctness}{
            \textbf{Return} $\mathcal{R}, A$\;
        }
        \Else{
            Remove the $R_i$ from $\mathcal{R}$; 
        }
    }
}
\textbf{Return} $\mathcal{R}, A$\;
\end{algorithm}

\xhdr{Reasoning trace evaluation.}
We consider the quality of the reasoning trace to be essential for the performance of \name. Evaluating the reasoning trace ensures both its reliability and correctness. This evaluation focuses on two main aspects: correctness and behavior.
For correctness, we examine the correctness of the answer, reasoning trace, and function calls. For answer correctness, in the case of multiple-choice questions, we compare the predicted option with the correct one. For open-ended reasoning questions, GPT-4 is prompted as a judge to determine if the prediction aligns with the correct answer.
For the reasoning trace, we use GPT-4 as a judge, with the question and the ground truth answer serving as references to assess the quality of the reasoning process.
For function calls, we verify that the correct tool is used, and we check the correctness of the argument names, argument value types, and the inclusion of any required arguments.
Even if the generated reasoning trace passes the correctness check, undesired behaviors may still occur, leading to incorrect reasoning during inference. In the behavior check, we examine issues such as hallucinations, arbitrary results, and repeated reasoning.
For hallucinations, we look for hallucinated placeholders in object names and IDs, such as drug names, disease names, and target IDs in function calls. Since IDs for drugs or diseases are not general knowledge, we eliminate reasoning traces where IDs appear without being shown earlier in the context.
The goal of \name is to generate verified reasoning traces, meaning the answers should be based on feedback from function calls. However, arbitrary results can arise when answers are derived from the model's unverified internal knowledge rather than tool feedback. We remove reasoning traces that are based on general knowledge instead of the feedback from the tools.
In more complex cases, \textsc{Solver} may generate reasoning traces that include repeated thoughts or function calls. 
For repeated thoughts, we assess the similarity between them, and for repeated function calls, we identify steps with identical function calls using the same arguments. 
We remove reasoning traces that involve repeated thoughts and function calls.
If the reasoning trace passes all checks, it is retained. However, if it fails due to errors in correctness or undesired behaviors, it is discarded. This evaluation ensures that only high-quality reasoning traces are considered in training data.

\subsection{Iterative Training for \toolrag}
\toolrag is used in both \name inference process and the training data collection phase.
It utilizes gte-Qwen2-1.5B-instruct~\cite{li2023towards} as the base model, which is fine-tuned on pairs of requirements and tool descriptions using the multiple negatives ranking loss.
In the \textsc{Tool Provider} module, we use the \toolrag to identify tools beyond the initial list retrieved from the reference information of the question. While the \toolrag requires training data from reasoning traces, we propose an iterative training process for \toolrag, where it is trained on the generated reasoning traces, which in turn helps improve the generation of future reasoning traces.
In the first stage, since the \toolrag is not yet available, we rely solely on the initial set of tools $\mathcal{\hat{P}}_0$ that are obtained from the reference information of the question, to generate the reasoning trace. From this trace, we extract pairs of tool requirements and tool descriptions, which are then used to train the \toolrag.
In the second stage, after the initial training of \toolrag, we use it to select tools instead of relying exclusively on $\mathcal{\hat{P}}_0$. This approach allows the reasoning trace to better reflect real-world use cases, as tools are now retrieved directly by the \toolrag. Using the data collected from this stage, we continue to gather new pairs for further training of the \toolrag.
This process is repeated iteratively, continually refining both the \toolrag and the quality of reasoning trace generation.

\section{Training \name model}
\label{sec:training_txagent}
To enable the multi-step reasoning and function call capabilities of \name, we fine-tune LLMs using the \trainset dataset designed to encompass the diverse behaviors required by \name.
This section introduces the training dataset \trainset and the training strategies.

\subsection{\name Training Dataset: \trainset dataset}
\label{sec:dataset_convert}
We use three agent systems to generate 
three training datasets, including a tooling dataset, a therapeutic question dataset, and a reasoning trace dataset.
The question dataset comprises 85,340 therapeutic questions, while the reasoning trace dataset includes 177,626 reasoning steps and 281,695 function calls.
Then, we begin by integrating the question with a reasoning trace and incorporating augmented tools. Next, we break down the complete reasoning trace into step-wise training data.
This process results in the creation of the \trainset dataset that contains 378,027 instruction tuning data samples.
The \trainset dataset is generated by randomly sampling from drugs in the FDA drug label database and disease phenotypes from the PrimeKG database.
To prevent any leakage of evaluation data through the training data, we remove all drugs approved after 2023.

\xhdr{Constructing step-wise training data.}
During supervised fine-tuning, in order to enable \name to have step-wise reasoning and function call abilities, we apply step-wise supervision on the thoughts and function calls at each reasoning step. Given a question $ Q $, a reasoning trace $ \mathcal{R} = \{ R_1, R_2, R_3, \dots, R_M \} $ consisting of $ M $ reasoning steps, and the final answer $ A $, where each reasoning step $ R_i $ is represented as a tuple $ R_i = \{T_i, \mathcal{C}_i, \mathcal{E}_i\}$ — with $ T_i $ and $ \mathcal{C}_i $ being the thought and function calls at the $ i $-th step, and $ \mathcal{E}_i $ results of function calls — we decompose the reasoning trace into $ M $ step-wise samples for supervision. Each of these step-wise samples consists of an input and an output for the fine-tuned LLM.

For each $ i \in \{1, 2, \dots, M-1\} $, the input to the model consists of the 
system prompt $S$, question $ Q $, a set of available tools at step $i$ denoted as $ \mathcal{P}_i $, and the reasoning trace up to the previous step, denoted as $ \mathcal{R}_{1:i-1} $, which represents the reasoning steps from $ R_1 $ to $ R_{i-1} $. 
The output of the model is the components $ [T_i, \mathcal{C}_i] $, which correspond to the thought and function calls at step $ i $.
At the final step $ M $, the input consists of the system prompt $S$, question $ Q $ and the reasoning trace up to the $ M-1 $-th step, i.e., $ \mathcal{R}_{1:M-1} = \{ R_1, R_2, \dots, R_{M-1} \} $, as well as the tools available up to that step, $ \mathcal{P}_M $. The output consists of the thought $ T_M $, the final function call to the \textsc{Finish} tool $ \mathcal{C}_M $, and the final answer $ A $.
Thus, for each $ i \in \{1, 2, \dots, M\} $, the $ i $-th step-wise sample is:

\begin{equation}
\begin{aligned}
\text{Input: } & \left[S, Q, \mathcal{R}_{1:i-1}, \mathcal{P}_i \right], \quad \text{Output: } \left[ T_i, \mathcal{C}_i \right] \quad \text{for} \quad i \in \{1, 2, \dots, M-1\}, \\
\text{Input: } & \left[S, Q, \mathcal{R}_{1:M-1}, \mathcal{P}_M \right], \quad \text{Output: } \left[ T_M, \mathcal{C}_M, A \right] \quad \text{for} \quad i = M.
\end{aligned}
\end{equation}
While each step in the reasoning trace may involve multiple function calls, we introduce an argument, \textsc{id}, which is a randomly generated string, to uniquely identify each function call. This \textsc{id} is added to both the function call arguments $\mathcal{C}_{i,k}$, and the corresponding results returned by the function $\mathcal{E}_{i,k}$.
The \textsc{id} is added to the input reasoning trace $\mathcal{R}_{1:M-1}$
and is removed in model output as it's random unpredictable string.

This step-wise decomposition allows for effective supervision of the reasoning process, with each sample providing contextual information about the model's reasoning at every intermediate step. At the final step, both the reasoning components and the final answer are output together, marking the completion of the reasoning process.

\subsection{Training data augmentation}
We design several training data augmentation strategies to ensure that \name is trained to perform function calls based on contextual information and can generalize to new tools.

\xhdr{Augmenting tools.}
To prevent over-fitting to the tools in \toolbox, we apply augmentation to the tool descriptions by randomly rephrasing all fields of a tool. 
For each tool in \toolbox, we prompt the LLM to rewrite the original description and generate 20 distinct versions of the tool's name, function description, argument names, and argument descriptions.
For each training sample in \trainset, we randomly select from these rewritten fields to create a new, augmented tool description. Then, we replace the tool name and argument names in the function call arguments to generate the augmented training sample.
This strategy enables \name to learn how to call functions based on the tool names and arguments, rather than memorizing the specific functions encountered during training. 
As a result, the model can generalize to new and unseen tools during inference, enabling flexible scaling of \toolbox.

\xhdr{Extending the available tool set.}  
The available tool set $\mathcal{P}$ (index note of $i$-th step is omitted here for simplicity) in each sample of \trainset consists of tools used in the reasoning traces. However, during inference, the tools retrieved by the imperfect \toolrag may include additional candidates, making it challenging for \name to select the most suitable tools from the returned set.
To address this, we enhance the tool set $\mathcal{P}$ by including all tools retrieved by \toolrag, not just those explicitly used in the reasoning traces. Additionally, we randomly sample several tools from \toolbox and add them to $\mathcal{P}$.
This approach ensures that \name learns to effectively select the correct tool from a broader set of candidate tools.

\xhdr{Shuffling the tool list.}  
To mitigate any potential bias introduced by the position of tools in the tool list $\mathcal{P}$, we shuffle the tools in $\mathcal{P}$. This ensures that the order in which tools appear does not influence the ability of \name to select the correct tool. By randomizing the positions of the tools, we encourage \name to focus on the context and descriptions of the tools, rather than their position in the list.
This strategy helps the model learn to make tool selections based on the contextual information rather than relying on the order of the tools in the tool set.

\xhdr{Replacing long results to results summary.}
To ensure the training data fits within the context window while preserving the overall reasoning trace, we shorten samples that exceed the maximum context limit by replacing the full tool results with summarized versions. This process begins with the earliest step in the reasoning trace and continues until the total length of the sample is within the context limit.

\subsection{Training design}
\label{sec:training_design}

\xhdr{Model.}
We use pre-trained LLMs, such as Llama3.1-8B-Instruct, as the base models for fine-tuning on the \trainset. The Llama3 series is built on the Transformer architecture~\cite{vaswani2017attention}, which leverages self-attention mechanisms to process input sequences in parallel. This enables efficient learning of contextual relationships between tokens.
Our model initialization begins with loading the pre-trained weights from the instruction-tuned version of the Llama3 series, specifically fine-tuned on question-and-answer data. To further adapt the model to our specific task, we apply Low-Rank Adaptation (LoRA) fine-tuning~\cite{hu2021lora}. LoRA enhances the fine-tuning process by introducing low-rank updates to the pre-trained weights, reducing computational costs and the number of parameters trained. This allows us to efficiently fine-tune the model while preserving the knowledge from the pre-trained LLM.

\xhdr{Training process.}
During the training process, the input of one training sample  
$\left[S, Q, \mathcal{R}_{1:i-1}, \mathcal{P}_{1:i-1} \right]$
and the corresponding output $\left[ T_i, \mathcal{C}_i \right]$
are formatted according to the instruction-following format of the LLM (e.g., the chat template of Llama).
The formatted text is then processed by the LLM tokenizer to generate a sequence of tokens $\mathbf{x} = \{x_1, x_2, \dots, x_N\}$,
where $N$ is the total number of tokens. These tokens are embedded and passed through the model for autoregressive prediction. The model predicts the next token $x_t$ based on all previously generated tokens $\{x_1, x_2, \dots, x_{t-1}\}$, producing a conditional probability distribution $p(x_t \mid x_1, x_2, \dots, x_{t-1})$.
The training objective is to minimize the autoregressive loss, but only for the tokens corresponding to the output sequence $ \mathbf{x} \subseteq \mathbf{x}_\text{out} $. The loss is defined as:
\begin{equation}
\mathcal{L} = - \sum_{t \in \text{Idx}_\text{out}} \log p(x_t \mid x_1, x_2, \dots, x_{t-1}),
\end{equation}
where $ \text{Idx}_\text{out} $ represents the indices of tokens in $ \mathbf{x} $ corresponding to the output sequence $ \mathbf{x}_\text{out} $. By focusing on the output tokens, this loss ensures that the model learns to generate thought and function calls instead of over-fitting to the results from tools.

\subsection{Training implementation}
\xhdr{Training resources.}
We use the Nvidia H100 GPU cluster provided by the Kempner Institute for the Study of Natural and Artificial Intelligence at Harvard University to train \name.
For training the \name-8B model, we utilize 4 GPUs, totaling 320GB of GPU memory.
Training \name-8B requires 9.93 GPU days.

\xhdr{Training infrastructure.}
The training infrastructure of \name is modified based on several key libraries, including TRL~\cite{vonwerra2022trl}, Alignment Handbook~\cite{Tunstall_The_Alignment_Handbook}, Transformers~\cite{wolf-etal-2020-transformers}, Deepspeed~\cite{rasley2020deepspeed}, and PyTorch~\cite{paszke2017automatic}. The fully sharded data parallel (FSDP) technique is employed as the multi-GPU distributed training method. In this setup, the model's parameters are split across multiple GPUs to reduce memory usage, enabling the training of \name with large backend LLMs and long context windows. This approach efficiently distributes both computation and model weights, allowing for better scalability while minimizing communication overhead.
For multi-node training, we leverage the PyTorch implementation of FSDP. For single-node training, the Deepspeed implementation of FSDP is utilized, which is also referred to as Deepspeed Stage 3.

\section{Benchmarking \name}
\label{sec:benchmark_details}

\subsection{Benchmarks}
We constructed five evaluation benchmarks, including DrugPC, BrandPC, GenericPC, DescriptionPC, and TreatmentPC.
Given that LLMs have been pretrained on vast amounts of publicly available internet data, there is a risk of potential data leakage, meaning that LLMs may have previously encountered similar questions. To mitigate this risk in our evaluation datasets, we focused on creating new datasets centered around drugs approved by the FDA in 2024, reducing the likelihood that the LLMs have been exposed to this specific information. Statistics of all benchmarks are shown in Table~\ref{table:fda_11tasks}.

\xhdr{DrugPC: A comprehensive benchmark covering 11 common therapeutic tasks.} We created the DrugPC dataset, which includes 3,168 questions covering 11 common tasks related to therapy. These sub-tasks include drug overview, drug ingredients, drug warnings and safety, drug dependence and abuse, dosage and administration, use in specific populations, pharmacology, clinical information, nonclinical toxicology, patient-focused information, and storage and supply (as detailed in Table~\ref{table:fda_11tasks}).
To facilitate evaluation, the dataset is formatted as multiple-choice questions, with each question followed by several options (most having 4 options, with some having 2 or 5). The dataset construction process follows these steps:
1) We classify the sections within FDA documents and map them to 11 tasks. The specific fields for each task are outlined in Table~\ref{table:fda_11tasks}.
2) For question construction, we use the text from relevant sections of FDA documents as context. Using the question construction multi-agent system \questgen, we create questions, multiple-choice options, and corresponding answers that can be answered using the provided context.
The evaluation process checks whether the questions are answerable based on the context provided and ensures that the answers are accurate according to the given information.
3) After construction, a human evaluation process is conducted to carefully review and refine the questions and answers, ensuring that non-biomedical content, such as information about drug manufacturers, is excluded.

\xhdr{BrandPC/GenericPC: Datasets representing drug name in brand and generic forms.} LLM-based methods have been shown to be sensitive to variations, such as representing drugs by either their brand or generic names. To assess the robustness of \name, we transform the DrugPC dataset into two versions: BrandPC and GenericPC. In these versions, drug names are systematically replaced with their respective brand or generic names. Problems that do not involve drug names in the questions or options remain unchanged, while those requiring conversion between brand and generic names are also kept as is.

\xhdr{DescriptionPC: A benchmark representing drugs with detailed descriptions.} 
The drug name plays a crucial role in enabling LLM-based methods to effectively answer questions. However, to evaluate the models' generalization capabilities in the absence of explicit drug names, we introduce the DescriptionPC benchmark. In this benchmark, drug names are replaced with detailed descriptions that include information such as indications, mechanisms of action, contraindications, and drug interactions. 
To ensure the validity of the dataset, we manually remove questions in the DrugPC benchmark that cannot be answered after replacing the drug name with its description. This process results in 626 questions, forming the DescriptionPC benchmark.
While a model might infer an answer without explicitly identifying the drug name from its description, ensuring that the prediction is based on the correct drug rather than exploiting patterns is critical. 
To address this, the DescriptionPC benchmark incorporates a two-step evaluation process: drug identification and answer correctness evaluation.
\textbf{Drug identification}: The model must identify the drug name based on the provided description. Since multiple drugs can share similar descriptions, we construct the ground truth for this step by first collecting drug descriptions corresponding to their original names. We then identify similar drugs that can be described in the same way and include them in the ground truth.
\textbf{Answer correctness evaluation}: Using the drug names predicted in the first step, the model is tasked with selecting the correct answer from multiple-choice questions.
During the two-step evaluation process, if the drug identification in the first step is incorrect, the second step is automatically marked as incorrect, regardless of the answer’s correctness in that step. This approach ensures that the evaluation rigorously tests the model’s reasoning based on the intended drug descriptions.

\xhdr{TreatmentPC: A specialized treatment benchmark for precision therapy in targeted conditions.}
While multiple indications can be applied to a single disease, patients with specific conditions, such as pregnancy or comorbidities, require specialized treatment approaches, such as customized drug selection and dosage adjustments.
The TreatmentPC benchmark is designed to address such specialized treatment scenarios by generating questions based on the varying application conditions of drugs. This is achieved using the question construction system \questgen.
We first select drugs approved by the FDA in 2024, identifying their indicated diseases. 
For each disease, we compile all associated treatments and analyze the unique attributes of each drug. 
This analysis is conducted by examining FDA documents, including information on indications, usage in specific populations, safety warnings, precautions, and contraindications.
Next, we generate multi-choice questions that specifically account for differences among drugs. The answer options represent treatments for the disease, but only one is suitable based on the patient’s specific condition. For instance, we include scenarios where a patient is taking other medications that are contraindicated for certain treatments.
The TreatmentPC benchmark requires the model to do a thorough analysis of the patient’s condition before determining an appropriate solution.

\subsection{Evaluation strategy}
\label{sec:eval_strategy}
To assess performance on the aforementioned benchmarks, we employ two evaluation strategies: multiple-choice evaluation and open-ended evaluation. Examples of both multiple-choice and open-ended questions can be found in Table~\ref{tab:example_question_type}.

\xhdr{Multi-choice evaluation.} In this approach, the question is accompanied by multiple options, and the model must select the correct answer from these options. Accuracy across the dataset is reported as the evaluation metric.

\xhdr{Open-ended evaluation.} The model is presented with only the question, without any options, and is required to generate an open-ended answer. Evaluating such answers is inherently challenging. To address this, we introduce an additional step: the generated open-ended answer is provided as context, and the model is tasked with selecting the correct answer from multiple options based on this context. This allows for the evaluation of open-ended questions by producing quantitative results.

\xhdr{Performance metrics.}
For both multi-choice evaluation and open-ended evaluation, we report the accuracy on the benchmark dataset as the performance metrics.

\section{Settings for analysis of \name}

\subsection{\toolbox vs. LLM-as-tools}
\label{sec:llm_as_tool}
In the experiments comparing \toolbox with LLM-as-tools, we prompt the LLM with the following instruction to make it function as tools:
\begin{promptsllm}[LLM-as-a-tool]
You are a function that answers the questions based on your given description and given input. Do not answer questions that you don't have knowledge about.
\par
Here is your definition: {\{tool description\}}.
\par
Here is the input to the function:{\{function call arguments\}}. 
\par
The tool response:
\end{promptsllm}
In this instruction, the function call arguments generated by \name serve as the input, while the tool description obtained from the \toolbox is used as a reference. The LLM is then prompted to simulate the tool's outputs.

\subsection{Limit \name to function calls only, no thoughts}
\label{sec:no_thought_inference}
To verify the role of reasoning thoughts in \name, we build a modified version of \name that does not generate reasoning thoughts (Algorithm~\ref{alg:txagent_inference_nothought}).
This modified version of \name follows a multi-step inference process where, at each step, instead of explicitly reasoning through generating intermediate thoughts, the model directly produces function calls. The process starts with the initialization of the reasoning trace $\mathcal{R} \gets \{\}$, the set of available tools $\mathcal{P} \gets \mathcal{P}_0$, and a step counter $i \gets 0$. In each iteration, the model generates function calls or the final answer:
\begin{equation}
U_i = \mathcal{F}_{TX}(Q, \mathcal{R}_{i-1}, \mathcal{P}_i).
\end{equation}
If $U_i$ contains textual content, it is assigned as the final answer $A$, and the \textsc{Finish} tool is executed to terminate the reasoning process, returning $\mathcal{R}_i$ and  $A$. Otherwise, $U_i$ is the function calls arguments $\mathcal{C}_i$, which are executed. If $\mathcal{C}_i$ includes a call to \textsc{ToolRAG}, the available tools $\mathcal{P}_i$ are updated accordingly. The reasoning trace is iteratively updated as:
\begin{equation}
\mathcal{R}_i \gets \mathcal{R}_{i-1} \cup \{\mathcal{C}_i, \mathcal{E}_i\}.
\end{equation}

\begin{algorithm}[H]
\caption{\name multi-step inference process without thoughts.}
\label{alg:txagent_inference_nothought}
\KwIn{Question $Q$, \toolbox $\mathcal{B}$, Initial available tools $\mathcal{P}_0$}
\KwOut{Reasoning trace $\mathcal{R}$, final answer $A$}

Initialize $\mathcal{R} \gets \{\}$, tools $\mathcal{P} \gets \mathcal{P}_0$, step $i \gets 0$ \;

\While{Reasoning is incomplete}{
    $i \gets i + 1$\;

    Generate function calls or final answer: $U_i = \mathcal{F}_{TX}(Q, \mathcal{R}_{i-1}, T_i, \mathcal{P}_i)$
    
    \If{$U_i$ contains text}{
        Split the text as the final answer: $A$; \\
        Execute \textsc{Finish} tool to end the multi-step reasoning; \\
        \textbf{Return} $\mathcal{R}_i, A$\;
    }
    \Else{
    $\mathcal{C}_i  = U_i$; \\
    \If{call to \textsc{ToolRAG} in $\mathcal{C}_i$}{
        Execute \textsc{ToolRAG} and update $\mathcal{P}_i$; } 
    \Else{
    Execute tools from $\mathcal{C}_i$;
    }
    Update reasoning trace: $\mathcal{R}_i \gets \mathcal{R}_{i-1} \cup \{\mathcal{C}_i, \mathcal{E}_i\}$\;
    }
}
\textbf{Return} $\mathcal{R}, A$\;
\end{algorithm}

\section{Prompt sketches}
This section shows the prompts used to build agents in the data generation multi-agent systems.
For simplicity, we provide prompt sketches that contain an outline of the full prompt that summarizes key points and omits unnecessary details.

\subsection{System prompt for \name}

\begin{promptsllm}[TxAgent]
You are a helpful assistant that will solve problems through detailed, step-by-step reasoning and actions based on your reasoning. Typically, your actions will use the provided functions. You have access to the following functions. {\{functions\}}
\end{promptsllm}

\subsection{Prompts for \toolgen}
\begin{promptsllm}[Summarizer]
{\{API schema\}}

\vspace{1em}

\noindent Using the provided {\{database name\}} API Schema, generate all possible specific functional commands in words with no code. Output them in a list.
\end{promptsllm}

\begin{promptsllm}[Tool Generator (for openFDA tools)]
You are a helpful assistant for generating functions based on the field descriptions and API schema of openFDA:\newline
{\{API schema, field descriptions, and example functions\}}\newline
Guidelines:
\begin{itemize}
    \item Generate \textbf{two functions}: one function retrieves the drug name based on the field information, and the other function retrieves information for that field based on drug names.
    \item Align the function with the expected fields and descriptions.
    \item Each function must be unique and different from existing examples.
    \item Fields should contain search\_fields and return\_fields:
    \begin{itemize}
        \item search\_fields is a \textbf{dict}, where the keys are the function input parameters and the values are the fields to be searched.
        \item return\_fields is a \textbf{list} of field names from which information must be returned.
    \end{itemize}
\end{itemize}
The capabilities of the functions should be related to the given capabilities: {\{capabilities\}}
\end{promptsllm}

\begin{promptsllm}[Tool Generator (for OpenTarget tools)]
You are a helpful assistant for generating functions based on the OpenTarget API schema:\newline
{\{API schema and example functions\}}\newline
Guidelines for the generated function:
\begin{itemize}
    \item The function should align with the schema's functional and structural requirements.
    \item The function's name, description, input parameters, and schema should be unique and different from the existing example functions.
    \item The function capabilities should be related to the given capabilities: {\{capabilities\}}
\end{itemize}
\end{promptsllm}

\begin{promptsllm}[Tool Checker]
You are a helpful assistant who generates test queries based on a given function. You are provided the following:
\begin{itemize}
    \item Function: {\{generated tool\}}
    \item Related keywords and information for questions and queries:{\{additional information\}}
\end{itemize}
Based on the provided function, you must generate {\{number\}} different questions in natural language that require using the function. 
\vspace{1em}

Guidelines:
\begin{itemize}
    \item The questions should be specific and diverse; avoid general questions
    \item Function calls must include ``name" and ``arguments" arguments
    \item Question examples: {\{examples\}}
\end{itemize}
\end{promptsllm}

\subsection{Prompt for \questgen}
\begin{promptsllm}[Information Extractor (for disease-centered personalized treatment questions)]
You are provided the following information:
\begin{itemize}
    \item \textbf{Disease Information:} These phenotypes or symptoms in the following disease-related information will be used to construct a patient profile. { \{disease desc info\}}
    \item \textbf{Paired Drug Information:} Here is a side-by-side comparison of multiple drug options that help in designing design patient conditions. Consider the side effects, drug interactions, contraindications, and other aspects of these drugs in deciding which patient-specific factors would require someone to take one drug instead of the other options. For example, one drug may be a better option than the others given specific adverse drug-drug interactions, warnings, age restrictions, patient population restrictions, pregnancy considerations, and contraindications. Include such factors in the constructed patient profile to make one drug the definitive correct answer. { \{drug information\}}
\end{itemize}

Generate a comparison analysis of the selected drugs based on the provided information. Show the differences between the drugs and provide evidence for the differences.
\end{promptsllm}
\begin{promptsllm}[Question Generator (for disease-centered personalized treatment questions)]
You are an assistant specializing in creating advanced biomedical multiple-choice questions focused on drug treatments given various patient-specific information like diseases, phenotypes, and genetic variation. 
\vspace{1em}

Guidelines:
\begin{itemize}
    \item Frame questions around patient case scenarios, where a patient is diagnosed with a disease or exhibits specific phenotypes, and the goal is to identify the most suitable treatment. You may also provide protein targets or genes. If additional info is given in the Personalized Information section below, incorporate this info into the profile being constructed.
    \item Construct questions and answer choices that compare multiple similar drug treatments and select the most suitable one given the patient's particular conditions. Incorrect answer choices could be drugs indicated for the disease but unsuitable for this particular patient due to factors like age, comorbidities, or dosage considerations. The correct answer should be the most appropriate drug for the patient’s specific profile.
    \item \textbf{Selected Tools:} Generate questions related to these functions. {\{selected tools\}}
    \item \textbf{Disease Information:} Use phenotypes or symptoms in the following disease-related information to construct the patient profile. { \{disease desc information\}}
    \item \textbf{Personalized Information:} When constructing the patient profile, use the following analysis of the side-by-side drug comparison. Consider the side effects, drug interactions, contraindications, and other aspects of these drugs in deciding which patient-specific factors would require someone to take one drug instead of the other options. For example, one drug may be a better option than the others given specific adverse drug-drug interactions, warnings, age restrictions, patient population restrictions, pregnancy considerations, and contraindications. Include such factors in the constructed patient profile to make one drug the definitive correct answer. 
    Drug information: { \{ drug information \}}
    Drug comparison analysis:  { \{side by side drug comparison from Information Extractor agent\}}

Generate a question, answer, and explanation according to this format:
{ \{format outline\}}
\end{itemize}
\end{promptsllm}

\begin{promptsllm}[Question Generator (for tool-chain-centered questions)]
You are a helpful assistant for generating expert-level biomedical questions. Based on the given functions, generate a single independent question that focuses on the given drug. The question should be specific, diverse, and framed in multiple ways, requiring the use of as many functions as possible. Do not write a long question; break up the question into multiple sentences if needed. Do not include details that a scientist, physician, or patient would not know (\textit{e.g.,} ontology IDs like MONDO, EFO, CHEMBL, Ensembl/ENS).
\vspace{1em}

Use only the following information:
\begin{enumerate}
    \item Functions that can retrieve information related to the drug: { \{tool descriptions in the sampled tool chain\}}
    \item Related information from functions: { \{information obtained from tools\}}
    \item Related information from PrimeKG interactions: { \{drug or disease related information from the PrimeKG knowledge graph\}}
\end{enumerate}

Generate a question, answer, and explanation according to this format:
{ \{format outline\}}
\end{promptsllm}

\begin{promptsllm}[Question Generator (for drug-centered questions)]
You are a helpful assistant to generate meaningful and challenging multi-choice questions for expert biomedical researchers. Formulate biomedical questions and generate answers using only the drug name and field information provided below:
\begin{itemize}
    \item Drug generic name: {\{generic name\}}
    \item Drug brand name: {\{brand name\}}
    \item Specific field of information for the drug (\textit{e.g.,} contraindications): {\{field information\}}
\end{itemize}
\vspace{1em}

Other guidelines:
\begin{itemize}
    \item Generate multiple, different questions to utilize all of the provided information. Make sure the questions do not overlap in content.
    \item Formulate questions that can be answered without needing additional information beyond the field information provided.
    \item Ask questions in different ways. Don't always start with ``What" and ``Which".
\end{itemize}

Generate a question, answer, and explanation according to this format:
{ \{format outline\}}

\end{promptsllm}

\subsection{Prompts for \tracegen}

\begin{promptsllm}[Helper]
    Please act as a helper to provide solution hints for the next step in solving the question. Give some suggestions about what to do next, but never give the final answer or information that directly leads to the final answer. Only provide hints for one reasoning step. 
    \vspace{1em}
    
    Also, make sure the user's final answer contains the correct answer. If not, let the user do self-reflection and continue reasoning until the correct answer is found.
\begin{itemize}
    \item Question:{\{question\}}
    \item Correct final answer:{\{answer\}}
    \item Explanation of correct answer:{\{explanation\}}
    \item Previous reasoning steps:{\{reasoning trace\}}
\end{itemize}
\end{promptsllm}

\begin{promptsllm}[Solver]
    You must fully understand and solve a question through reasoning and function calls.
    \vspace{1em}

    Guidelines:
    \begin{itemize}
        \item For each step, you must generate a reasoning thought and correct function call. If needed, call multiple functions. 
        \item If you think you have answered the question, thoroughly reflect on your reasoning to verify you have in fact answered the question. If not, continue reasoning. If so, call the `Finish' function and provide your final answer, which should be 1) comprehensive, 2)~explain how you arrived at the answer, and 3) why the answer addresses the question.
        \item If the result from the last function call is empty or not useful, you must continue reasoning and call ToolRAG (or simulate a virtual ToolRAG call) to retrieve more tools.
        \begin{itemize}
            \item If the tool you need is in the Function List below, you must retrieve them using a virtual ToolRAG call that simulates obtaining the tool through ToolRAG.
            \item If the tool you need is not in the Function List below, you need to call ToolRAG.
            \item { \{Description of ToolRAG and virtual ToolRAG tools\}}
        \end{itemize}
        \item Do not answer the question based on general knowledge. You must answer the question based on the information returned by the tools.
        \item If all previous solution attempts have failed, do not repeat the same thoughts and function calls. Instead, come up with new solution approaches.
    \end{itemize}

Function List: { \{available tools description of the initial set of tools $\mathcal{\hat{P}}_0$\}}
\vspace{1em}

For each reasoning step, respond in this JSON format: { \{reasoning step format\}}
\vspace{1em}

For the final step, respond in this JSON format, providing the final answer and a detailed explanation:
{ \{final reasoning step format\}}

Previous reasoning steps: { \{previous multi-step reasoning trace\}}

Hint for next step: { \{solution hint from Helper agent\}}
\end{promptsllm}

\clearpage

\section*{References}

{
\spacing{0.85}
\bibliographystyle{naturemag}
\bibliography{refs}
}

\end{document}